\documentclass{article}
\newcommand\mydel{\mathrel{\overset{\makebox[0pt]{\mbox{\normalfont\tiny\sffamily delta}}}{\, \approx \,}}}

\newcommand\mystein{\mathrel{\overset{\makebox[0pt]{\mbox{\normalfont\tiny\sffamily Stein}}}{\, = \,}}}

\usepackage{caption}
\usepackage{microtype}
\usepackage{graphicx}
\usepackage{subfigure}
\usepackage{booktabs} %

\usepackage{hyperref}

\usepackage[accepted]{icml2024}

\usepackage{amsmath}
\usepackage{amssymb}
\usepackage{mathtools}
\usepackage{amsthm}

\usepackage[capitalize,noabbrev]{cleveref}

\theoremstyle{plain}

\theoremstyle{definition}

\theoremstyle{remark}

\usepackage[textsize=tiny]{todonotes}

\usepackage{nicefrac}       %
\usepackage{microtype}      %
\PassOptionsToClass{usenames,dvipsnames}{xcolor}
\definecolor{TUblue}{RGB}{0,105,170}

\usepackage{comment}

\newcommand{\stepsize}{\beta}

\newcommand\cut[1]{}

\newcommand{\squishlist}{
   \begin{list}{$\bullet$}
    { \setlength{\itemsep}{0pt}      \setlength{\parsep}{3pt}
      \setlength{\topsep}{3pt}       \setlength{\partopsep}{0pt}
      \setlength{\leftmargin}{1.5em} \setlength{\labelwidth}{1em}
      \setlength{\labelsep}{0.5em} } }

\newcommand{\squishlisttwo}{
   \begin{list}{$\bullet$}
    { \setlength{\itemsep}{0pt}    \setlength{\parsep}{0pt}
      \setlength{\topsep}{0pt}     \setlength{\partopsep}{0pt}
      \setlength{\leftmargin}{2em} \setlength{\labelwidth}{1.5em}
      \setlength{\labelsep}{0.5em} } }

\newcommand{\squishend}{
    \end{list}  }

\newtheorem{claim}{Claim}{}
{}
{}
{}
{}

\newcommand{\half}{\mbox{$\frac{1}{2}$}}

\newcommand{\real}{\mbox{$\mathbb{R}$}}

\newcommand{\gauss}{\mbox{${\cal N}$}}

\newcommand{\myvec}[1]{\mbox{$\mathbf{#1}$}}
\newcommand{\myvecsym}[1]{\mbox{$\boldsymbol{#1}$}}

\newcommand{\vdelta}{\mbox{$\myvecsym{\delta}$}}

\newcommand{\veta}{\mbox{$\myvecsym{\eta}$}}

\newcommand{\vmu}{\mbox{$\myvecsym{\mu}$}}
\newcommand{\vlambda}{\mbox{$\myvecsym{\lambda}$}}

\newcommand{\vSigma}{\mbox{$\myvecsym{\Sigma}$}}

\newcommand{\vtau}{\mbox{$\myvecsym{\tau}$}}

\newcommand{\vd}{\mbox{$\myvec{d}$}}

\newcommand{\vg}{\mbox{$\myvec{g}$}}
\newcommand{\vh}{\mbox{$\myvec{h}$}}

\newcommand{\vm}{\mbox{$\myvec{m}$}}
\newcommand{\vn}{\mbox{$\myvec{n}$}}

\newcommand{\vv}{\mbox{$\myvec{v}$}}
\newcommand{\vw}{\mbox{$\myvec{w}$}}

\newcommand{\vx}{\mbox{$\myvec{x}$}}

\newcommand{\vA}{\mbox{$\myvec{A}$}}
\newcommand{\vB}{\mbox{$\myvec{B}$}}
\newcommand{\vC}{\mbox{$\myvec{C}$}}
\newcommand{\vD}{\mbox{$\myvec{D}$}}

\newcommand{\vF}{\mbox{$\myvec{F}$}}
\newcommand{\vG}{\mbox{$\myvec{G}$}}

\newcommand{\vI}{\mbox{$\myvec{I}$}}

\newcommand{\vK}{\mbox{$\myvec{K}$}}
\newcommand{\vL}{\mbox{$\myvec{L}$}}
\newcommand{\vM}{\mbox{$\myvec{M}$}}
\newcommand{\vN}{\mbox{$\myvec{N}$}}

\newcommand{\vQ}{\mbox{$\myvec{Q}$}}

\newcommand{\vS}{\mbox{$\myvec{S}$}}

\newcommand{\vU}{\mbox{$\myvec{U}$}}
\newcommand{\vV}{\mbox{$\myvec{V}$}}
\newcommand{\vW}{\mbox{$\myvec{W}$}}

\newcommand{\diag}{\mbox{$\mbox{diag}$}}

\newcommand{\be}{\begin{equation}}
\newcommand{\ee}{\end{equation}}
\newcommand{\bea}{\begin{eqnarray}}
\newcommand{\eea}{\end{eqnarray}}
\newcommand{\beaa}{\begin{eqnarray*}}
\newcommand{\eeaa}{\end{eqnarray*}}

\begin{document}

\newcommand{\ourshorttitle}{
Spectral-factorized Positive-definite Curvature Learning for NN Training
}
\newcommand{\ourtitle}{
Spectral-factorized Positive-definite Curvature Learning for NN Training
}
\icmltitlerunning{\ourshorttitle}

\twocolumn[

\icmltitle{\ourtitle}

\begin{icmlauthorlist}
\icmlauthor{Wu Lin}{vector}
\icmlauthor{Felix Dangel}{vector}
\icmlauthor{Runa Eschenhagen}{cambridge}
\icmlauthor{Juhan Bae}{vector,ut}
\icmlauthor{Richard E. Turner}{cambridge}
\icmlauthor{
Roger B. Grosse
}{vector,ut}
\end{icmlauthorlist}

\icmlaffiliation{vector}{Vector Institute, Canada}
\icmlaffiliation{cambridge}{Cambridge University, United Kingdom}
\icmlaffiliation{ut}{University of Toronto, Canada}

\icmlcorrespondingauthor{Wu Lin}{yorker.lin@gmail.com \vspace{-0.2cm}}

\icmlkeywords{Natural Gradient Descent, Second Order Method, Optimization, Deep Learning}

\vskip 0.3in
]

\printAffiliationsAndNotice{}  %

\begin{abstract}
Many training methods, such as Adam(W) and Shampoo, learn a positive-definite curvature matrix and apply an inverse root before preconditioning. Recently, non-diagonal training methods, such as Shampoo, have gained significant attention; however, they remain computationally inefficient and are limited to specific types of curvature information due to the costly matrix root computation via matrix decomposition.
To address this, we propose a Riemannian optimization approach that dynamically adapts spectral-factorized positive-definite curvature estimates, enabling the efficient application of arbitrary matrix roots and generic curvature learning.
We demonstrate the efficacy and versatility of our approach in positive-definite matrix optimization and covariance adaptation for gradient-free optimization, as well as its efficiency in curvature learning for neural net training.

\vspace{-0.2cm}
\end{abstract}

\vspace{-0.65cm}
\section{Introduction}
\vspace{-0.2cm}
Symmetric positive-definite (SPD) curvature learning is useful for designing training methods. 
Given the neural net (NN) weights $\vmu$, the gradient $\vg$ and a SPD curvature estimate $\vS \succ 0$, many training methods apply $\smash{\vmu \leftarrow \vmu - \beta_1 \vS^{-1/p}}\vg$ using a step size $\beta_1$ and subjecting the curvature estimation to a fractional $p$-root before inversion.
For example, adaptive methods like 
Adam(W)~\citep{kingma2014adam,loshchilov2017decoupled}  employ the gradient outer product~\citep[GOP,][]{duchi2011adaptive,kingma2014adam,agarwal2019efficient} to estimate a diagonal matrix $\vS$ and apply a $2$-root (i.e., $p=2$) before preconditioning.
Other well-known methods, like natural-gradient methods, use $\vS$ to estimate the Fisher information   matrix~\citep{amari1998natural,roux2007topmoumoute} as another type of curvature and apply a $1$-root  before preconditioning.
These methods highlight the need for an SPD curvature learning scheme for $\vS$ that supports generic curvature information and root computation.

Existing works \citep{martens2015optimizing,zhang2018noisy,anil2020scalable,shi2023distributed,lincan2024,vyas2024soap} have demonstrated the great potential of Kronecker-based non-diagonal methods for training large neural networks because using Kronecker structures significantly reduces the memory consumption of non-diagonal methods.
However, applying fractional roots for Kronecker-based methods such as Shampoo~\citep{gupta18shampoo,anil2020scalable,shi2023distributed} and its variants \citep{vyas2024soap} remains computationally and numerically challenging. 
This is because computing a \emph{matrix} fractional root is computationally intensive and must be done in high precision to avoid numerical instabilities~\citep{anil2020scalable,shi2023distributed}, preventing those methods from using fast, low-precision arithmetic~\citep{micikevicius2017mixed}. 
Because of this bottleneck, non-diagonal methods are often limited to
a specific case of the root and particular types of curvature information.
Making the matrix root computation for generic curvature learning fast and stable not only enables non-diagonal methods for modern low-precision training but also facilitates the investigation of new non-diagonal methods.
Thus, it is essential to have a flexible and efficient approach that can (i) apply arbitrary fractional roots, (ii) circumvent the numerical instabilities of the root computation, and (iii) support generic curvature information. 

We address this instability and inefficiency and present an update scheme to directly and dynamically adapt the spectral factorization $\smash{\vB \mathrm{Diag}(\vd) \vB^\top}$ of a SPD estimation $\vS$ for generic curvature information. 
We call this parameterization a \emph{spectral parameterization}, due to its connection to the spectral decomposition of symmetric matrices. 
Thanks to this parametrization,  we can efficiently apply any matrix fractional root to $\vS$ through elementwise operations on the eigenvalues $\vd$.
Our approach directly adapts eigenfactors and maintains the factorization without performing matrix decomposition.  
This makes our scheme amenable to running in low precision because we do not rely on unstable matrix decomposition algorithms.
However, directly optimizing the factors involves nontrivial constraints (i.e., $\vB$ is orthogonal and $\vd$ is positive) that must be dealt with.

We overcome challenges and make these contributions:
\vspace{-0.3cm}
\begin{itemize}

\item
We propose an update scheme to dynamically adapt the spectral parameterization \scalebox{0.8}{ $\vB \mathrm{Diag}(\vd) \vB^\top$} of a SPD matrix $\vS$ for generic curvature information by extending the Riemannian idea
\citep{glasmachers2010exponential,lin2023simplifying,lincan2024} for estimating $\vS$.
Surprisingly, 
our scheme coincides with both diagonal and full-matrix AdaGrad and RMSprop up to first-order (Claim \ref{claim:eigen_full_mat_case} and Sec.~\ref{sec:diag_case_rgd}) when using the GOP as curvature information.

\item 
We then extend our scheme to Kronecker-structured spectral factorizations, i.e.\,\scalebox{0.8}{$\vS = (\vS^{(C)} \otimes \vS^{(K)})$}, where $\otimes$ denotes a Kronecker product, and \scalebox{0.8}{$\vS^{(l)}:=  \vB^{(l)} \mathrm{Diag}(\vd^{(l)}) {\vB^{(l)}}^\top$} for \scalebox{0.8}{ $l \in \{C,K\}$}, which are crucial to scale the approach.
The Kronecker structure introduces
additional challenges, which we resolve by introducing an improved factorization
\scalebox{0.8}{$\vS = \alpha (\vS^{(C)} \otimes \vS^{(K)})$} with
constraints \scalebox{0.8}{ $\det(\mathrm{Diag}(\vd^{(l)})) = 1$}.

\item 
Empirically, we demonstrate the efficacy and versatility of our approach for generic curvature learning in SPD matrix optimization \citep{absil2009optimization} and gradient-free optimization \citep{wierstra2008natural}, and showcase its efficiency and numerical stability in low-precision NN training.
Our experiments highlight the effectiveness of our scheme as a spectral-factorized adaptive method for NN training (See Fig.~\ref{fig:results}).

\end{itemize}

\vspace{-0.5cm}
\section{Background}
\label{sec:background}
\vspace{-0.15cm}
To train an NN model,
we solve an unconstrained optimization problem. The objective function of the problem is often expressed as a sum of cost functions with $N$ observations:

\vspace{-0.35cm}
\resizebox{\linewidth}{!}{
\begin{minipage}{1.1\linewidth}
\begin{align}
    \textstyle
    \min_{\mu }\, \ell (\vmu):=\sum_{i=1}^{N} c(f(\vx_i; \vmu), y_i),
    \label{eq:org_opt}
\end{align}
\end{minipage}
}

where $\vx_i$ and $y_i$ are features and a label for the $i$-th observation, respectively,
$f(\cdot;\vmu)$ is an NN with learnable weights $\vmu$,  and $c(\cdot, y_i)$ is a cost function (e.g., cross-entropy loss) to measure the mismatch between the NN's output and $y_i$.

We consider adaptive methods to solve this problem, where we estimate a preconditioning matrix using a GOP. For many well-known adaptive methods such as RMSprop \citep{tieleman2012rmsprop} (see Eq.~\eqref{eq:rmsprop} with  $\gamma=1$) and AdaGrad \citep{duchi2011adaptive} (see  Eq.~\eqref{eq:rmsprop} with  $\gamma=0$), a square root (i.e., $p=2$) is introduced before inversion.

\vspace{-0.4cm}
\resizebox{\linewidth}{!}{
\begin{minipage}{1.1\linewidth}
\begin{align}
\text{Diagonal}:  \vS &  \leftarrow  (1-\stepsize_2\gamma ) \vS + \stepsize_2 \mathrm{Diag} (\mathrm{diag}(\vg\vg^T)),\nonumber \\
   \vmu & \leftarrow \vmu - \stepsize_1 \vS^{-1/p} \vg, \label{eq:rmsprop}
\end{align}
\end{minipage}
}

\vspace{-0.05cm}
where 
$\diag(\cdot)$ returns a vector that contains diagonal entries
of its input,  $\mathrm{Diag}(\cdot)$ turns its input into a diagonal matrix,
$\vS$ is a diagonal matrix, $\vmu$ is the learnable weight vector,  $\vg=\nabla_\mu \ell$ is a gradient vector, and $\vg\vg^T$ is the GOP. We often estimate $\vg$ using a mini-batch of observations.
Other works improve the performance of adaptive methods on CNNs using other roots of $\vS$, for instance $p=4$ in \citet{chen2021closing} and $p=1$ in \citet{lincan2024}.
\citet{duchi2011adaptive} consider a full-matrix version of AdaGrad (i.e., $p=2$) update scheme while \citet{lincan2024} consider a variant of  AdaGrad with a $1$-root (i.e., $p=1$).

\vspace{-0.4cm}
\resizebox{\linewidth}{!}{
\begin{minipage}{1.1\linewidth}
\begin{align}
  \text{Full-matrix}:  \vS &\leftarrow  (1-\beta_2\gamma ) \vS + \beta_2 \vg\vg^T=  \vS - \beta_2 (\gamma \vS-\vg\vg^T),\nonumber\\
   \vmu &\leftarrow \vmu - \beta_1 \vS^{-1/p} \vg, \label{eq:root_free}
\end{align}
\end{minipage}
}
where a fractional matrix $p$-root is needed.

Kronecker-based adaptive methods, like Shampoo \citep{gupta18shampoo}, 
also include a matrix root (e.g, $p=4$) in their update rule to update matrix weights.

{\bf Challenges of Computing Matrix Roots}\,
Computing a matrix root (e.g., a square root, $p=2$) requires matrix decomposition, such as eigendecomposition. As a numerical linear algebra algorithm, matrix decomposition can be numerically unstable in \emph{discrete, finite-precision} arithmetic. Matrix decomposition is slow in single precision and unstable in half precision.
Using low-precision data types \citep{micikevicius2017mixed} is essential for training large NNs because it boosts training speed and lowers memory consumption.
Existing methods like Shampoo \citep{gupta18shampoo,shi2023distributed} and its variants \citep{vyas2024soap} often perform the matrix decomposition in single (rather than half) precision and ameliorate the cost by performing the decomposition less frequently. However, an infrequent update scheme can significantly degrade the curvature estimation  \citep{vyas2024soap}, often necessitating specific mitigations. For example, Shampoo often uses a pre-tuned Adam(W) to stabilize its infrequent update \citep{shi2023distributed,vyas2024soap} via grafting \citep{agarwallearning}. 
Iterative methods for the root computation \citep{anil2020scalable} also have been proposed
for single precision but have not yet been used in half-precision due to numerical instability \citep{shi2023distributed}.
For the Shampoo-based curvature information, 
\citet{muon2024} propose the Newton-Schulz iteration based 
on spectral norm descent \citep{bernstein2024old}.
However, this method is limited to a specific matrix root (i.e., $p=4$) and is not easy to generalize to curvature information other than the Shampoo-based curvature.

\vspace{-0.25cm}
\subsection{Riemannian Idea for Generic Curvature Learning}
\label{sec:rgd_approach}
\vspace{-0.1cm}
Although recent non-diagonal methods primarily focus on the Shampoo-based approximation, it is equally important to explore other types of curvature information to develop new non-diagonal methods. 
Inspired by \citet{lincan2024}, we will propose an efficient update scheme that supports generic curvature information and arbitrary matrix roots.

We first review the approach  of \citet{lincan2024}. They propose an update scheme for generic SPD  curvature estimation of $\vS$. %
When using the GOP as curvature information, they show that the scheme in \eqref{eq:root_free} with $p=1$ is a simplified version of Riemannian gradient descent (RGD) on a Gaussian manifold \citep{amari2016information}, where $\vmu$ and $\vS$ in \eqref{eq:root_free} become Gaussian's mean and inverse covariance, respectively.
Concretely, they consider a Gaussian approximation problem and use the following procedure to obtain the scheme in \eqref{eq:root_free}.

{\bf Step 1}\, They first reformulate the original problem in \eqref{eq:org_opt} as a variational Gaussian approximation problem:

\vspace{-0.35cm}
\resizebox{\linewidth}{!}{
  \begin{minipage}{1.1\linewidth}
\begin{align}
    \min_{\mu, S \succ 0}\, \mathcal{L} (\vmu,\vS):= E_{w \sim q(w;\mu,S)}[ \ell(\vw) ] - \gamma \mathcal{Q}_q,
    \label{eq:ref_opt}
\end{align}
\end{minipage}
}

\vspace{-0.25cm}
where $\gamma \in \{0,1\}$ is defined in Eq.~\eqref{eq:root_free}, $\ell(\cdot)$ is the loss function in Eq.~\eqref{eq:org_opt},  a new symbol $\vw$ is used to denote the weights of the NN   because they are no longer learnable,  $q(\vw;\vmu,\vS)$ is a Gaussian with mean $\vmu$ and covariance $\vS^{-1}$, and \scalebox{0.8}{$\mathcal{Q}_q:=E_{w \sim q}[-\log q(\vw;\vmu,\vS)]=-\half \log \mathrm{det}(\vS)$} is the Gaussian's differential entropy.

{\bf Step 2}\, They then suggest performing RGD in this  parameter space  $\vtau:=\{\vmu,\vS\}$ of the Gaussian.

\vspace{-0.4cm}
\resizebox{\linewidth}{!}{
  \begin{minipage}{1.1\linewidth}
\begin{align}
    \mathrm{RGD}: \vtau \leftarrow \vtau - \stepsize [\vF_{\tau}]^{-1} \nabla_\tau \mathcal{L},\label{eq:rgd}
\end{align} 
\end{minipage}
}

\vspace{-0.1cm}
where \scalebox{0.8}{
$\vF_{\tau}:=E_{w\sim q}[\nabla_\tau \log q(\vw;\vtau) \nabla_\tau^\top \log q(\vw;\vtau)] \in \real^{ (l+l^2)\times (l+l^2)}$ } is the Fisher-Rao metric in coordinate $\vtau$  and  $l$ is the number of NN weights.
This metric is high-dimensional. The Moore-Penrose inverse is used 
in Eq.~\eqref{eq:rgd} if $\vF_{\tau}$ is singular.

{\bf Step 3}\, Simplifying the RGD step %

\vspace{-0.2cm}
\resizebox{0.9\linewidth}{!}{
  \begin{minipage}{1.1\linewidth}
\begin{align}
    \mathrm{RGD}: &\begin{bmatrix}
         \vS \\ \vmu
    \end{bmatrix}  \leftarrow
     \begin{bmatrix}
        \vS \\ \vmu
    \end{bmatrix}  - \stepsize
    \begin{bmatrix}
 -2 \frac{\partial S }{\partial S^{-1} } & \mathbf{0}\\
 \mathbf{0} & \vS^{-1} \\
    \end{bmatrix}
    \begin{bmatrix}
\partial_S \mathcal{L} \\
\partial_\mu \mathcal{L} 
    \end{bmatrix} \nonumber\\ 
     = &
      \begin{bmatrix}
     \vS +\stepsize (2\partial_{S^{-1}} \mathcal{L}) \\
        \vmu -\stepsize \vS^{-1} \partial_\mu \mathcal{L}
    \end{bmatrix}  \approx
     \begin{bmatrix}
     \vS -\stepsize ( \gamma\vS- \vg\vg^\top ) \\
        \vmu -\stepsize \vS^{-1}  \vg
    \end{bmatrix},
    \label{eq:rgd_ada}
\end{align}
\end{minipage}
}

\vspace{-0.1cm}
gives rise to the scheme in \eqref {eq:root_free} with \scalebox{0.8}{$p=1$}, where they use (i) the analytical inverse  metric
\scalebox{0.8}{
$[\vF_{\tau}]^{-1} = \begin{bmatrix}
 -2 \frac{\partial S }{\partial S^{-1} } & \mathbf{0} \\
 \mathbf{0} & \vS^{-1},
\end{bmatrix}$}, (ii) 
Stein's estimator for the Gaussian \citep{opper2009variational}, and (iii) the GOP as a Hessian approximation \citep{lincan2024} \scalebox{0.8}{$\nabla_\mu^2 \ell \approx   \vg\vg^T$}  
with a delta evaluation at mean $\vmu$:

\vspace{-0.5cm}
\resizebox{\linewidth}{!}{
  \begin{minipage}{0.92\linewidth}
\begin{align}
& 2\partial_{S^{-1}}\mathcal{L} \mystein E_{w\sim q}[\nabla_w^2 \ell ] -\gamma  \vS \mydel \nabla_\mu^2 \ell -\gamma\vS  \approx \vg\vg^T -\gamma\vS,\nonumber \\
&     \partial_\mu \mathcal{L} \mystein E_{w \sim q}[ \nabla_w \ell] \mydel \nabla_\mu \ell =\vg. 
\label{eq:delta_approx}
\end{align}
\end{minipage}
}

\vspace{-0.35cm}
\paragraph{
Other Types of Curvature Information
}
This approach supports other types of curvature information.
For example, the RGD step becomes Newton's method \citep{khan18a}  if setting \scalebox{0.8}{$\stepsize=1$} and \scalebox{0.8}{$\gamma=1$} and using the Hessian curvature  information as  \scalebox{0.8}{$2\partial_{S^{-1}}\mathcal{L} \approx \nabla_\mu^2 \ell - \vS$}.
\citet{lin2021tractable} shows that this step recovers covariance adaptation evolution strategy \citep{wierstra2008natural} for gradient-free optimization when setting \scalebox{0.8}{$\gamma=0$} and using the curvature information obtained  from the REINFORCE  \citep{williams1992simple} estimation.
When using curvature information obtained from 
Bayesian inference,
 SPD matrix optimization, and inverse problems, this update scheme corresponds to Kalman filtering  \citep{khan2023bayesian},
RGD \citep{lin2023simplifying},
and ensemble Kalman filtering \citep{chen2024efficient}, respectively.
This step also supports the reparametrization trick \citep{lin2020handling} and Gauss-Newton approximation \citep{osawa2019practical} 
as other types of curvature information for training Bayesian NNs.

\vspace{-0.35cm}
\paragraph{Reparametrization Invariance} 
The Gaussian reformulation reveals a Riemannian structure hidden in  Eq.~\eqref{eq:root_free} and provides a new way to learn a decomposition of $\vS$ on the fly by 
reparametrizing $\vS$ and exploiting the reparametrization invariance of RGD.
Claim~\ref{claim:linear_invariance}—proof in Appx.~\ref{app:proof_lemma_linear}—establishes the invariance for linear unconstrained reparametrization.

\begin{claim}
\label{claim:linear_invariance}
RGD is linearly invariant in smooth, unconstrained parameter spaces, including over-parametrized spaces. 
\end{claim}

\vspace{-0.2cm}

{\bf Limitations of Existing Methods}
Generally, it is unclear if 
the reparametrization invariance holds for nonlinear and constrained reparametrizations.
\citet{lincan2024} present a positive case for directly learning a (nonlinear) Cholesky factor of \scalebox{0.8}{$\vS^{-1/p}$} with \scalebox{0.8}{$p=1$} on the fly.
However, their approach is not easy to extend to other factional 
roots (i.e., \scalebox{0.8}{$p\neq 1$}). 
Other approaches, like \citet{khan18a,
lin2020handling,
lin2021tractable,
tran2021variational,tan2021analytic,godichon2024natural}, also have difficulty efficiently computing matrix roots without matrix decomposition.
Moreover, most works do not preserve or demonstrate the invariance.

{\bf Challenges of Learning Constrained  Parametrizations via RGD}\, 
Performing RGD in a constrained coordinate is nontrivial because we have to satisfy parameter/coordinate constraints while taking a Riemannian metric into account.
For example, consider learning a spectral parameterization of $\vS=\vB \mathrm{Diag}(\vd)\vB^T$, where $\vB$ is an orthogonal matrix and $\vd$ is a vector with positive entries.
The RGD step in Eq.~\eqref{eq:rgd} does not
guarantee that these constraints are satisfied.
Many Riemannian approaches do not apply to the Gaussian problem because they are limited to certain constraints and tied to specific metrics.
For example, existing methods \citep{li2020efficient,kong2022momentum} only consider the orthogonal constraint for canonical metrics \citep{tagare2011notes} studied in the Riemannian optimization literature.
On the other hand, the Gaussian problem induces a non-standard metric for the orthogonal matrix $\vB$ and the vector $\vd$ jointly. 
The metric is necessary as it allows us to leverage the reparameterization invariance, enabling the design of efficient and stable adaptive methods.
Other methods like retraction-based methods \citep{boumal2014manopt} either use matrix decomposition to handle constraints or stick to the canonical parametrization $\vS$ for the SPD constraint, which contradicts our goal of solving the  problem with any $p$-root efficiently and stably.

{\bf Challenges of Efficiently Computing Riemannian Gradients
and  Handling a Singular Metric
}
Recall that the simplification step (i.e., Step 3) turns a computationally expensive RGD step in Eq.~\eqref{eq:rgd} involving the high-dimensional metric inversion into a more efficient adaptive update scheme.
Without an analytical metric inversion, we cannot simplify or explicitly express the RGD step as an adaptive update scheme.
When changing coordinates, the metric representation has to be changed accordingly \citep{lee2018introduction}.
However, an over-parameterized reparametrization can lead to a singular metric and thus complicate the metric inversion. 
For example, the metric is changed 
from block-diagonal and invertible  \scalebox{0.8}{$\vF_\tau = E_{w\sim q}[\nabla_\tau \log q(\vw;\vtau)\nabla_\tau^\top \log q(\vw;\vtau)] $} to non-block-diagonal and singular  \scalebox{0.8}{$\vF_\eta = E_{w\sim q}[\nabla_\eta \log q(\vw;\veta) \nabla_\eta^\top \log q(\vw;\veta)] $}, where \scalebox{0.8}{$\vtau=\{\vmu,\vS\}$} is a canonical coordinate and \scalebox{0.8}{$\veta=\{\vmu,\vd,\vB\}$} is a spectral coordinate.
The singularity occurs when $\vd$ has repeated entries.
Consequently, simplifying the (Moore-Penrose) inverse of this non-block-diagonal and singular metric is nontrivial. 
In other words, 
the inverse metric no longer admits an \emph{explicit}, simple, and analytical form.
The simplification step is more challenging in Kronecker-factorized cases because ambiguity (see  Sec.~\ref{sec:mat_gauss_nn}) arising from Kronecker factorization renders the metric always singular and complicates the simplification step.
Other approaches, such as implicit computation via automatic differentiation (auto-diff) \citep{salimbeni2018natural}, can result in implicit and inefficient schemes for structured cases, 
as auto-diff fails to leverage structures in both Kronecker-based $\vS$ and GOP-structured curvature for efficient computation.

\vspace{-0.4cm}
\section{
Spectral-factorized Curvature Learning
}\label{sec:fast-fngd}

\vspace{-0.2cm}
Our goal is to propose training schemes that support
generic curvature information and
offer the flexibility to apply arbitrary matrix roots and other operations needed for other applications, such as log-det, at a low cost.
To do so, we propose directly learning $\vB$ and $\vd$ in a spectral parametrization of \scalebox{0.8}{ $\vS = \vB \mathrm{Diag}(\vd) \vB^{\top}$} \emph{without} storing $\vS$, where $\vB$ is an orthogonal square matrix and $\vd$ is a vector with positive entries.
Our learnable spectral factor allows us to easily compute any fractional root and inversion \scalebox{0.8}{ $\smash{\vS^{-\nicefrac{1}{p}}=\vB \mathrm{Diag}(\vd^{-\nicefrac{1}{p}}) \vB^\top}$} through elementwise roots on $\vd$, instead of matrix roots on $\vS$.
We then extend our approach to Kronecker cases to support large-scale applications. Our approach is efficient as updating $\vB$ and $\vd$ does not involve any matrix decomposition.

Our starting point is that the adaptive method in \eqref{eq:root_free} is the Riemannian solution to the Gaussian problem discussed in Sec.~\ref{sec:rgd_approach}.
Because RGD is invariant under coordinate transformations (see  Claims \ref{claim:linear_invariance}, \ref{claim:eigen_full_mat_case}), our idea is to change coordinates so that the Riemannian solution becomes an adaptive update rule for spectral coordinates.
We propose using local coordinates to overcome the challenges of using spectral-constrained coordinates (see  Sec.~\ref{sec:rgd_approach}).

{\bf Constraint Satisfaction and Metric Diagonalization  via \emph{Local} Coordinate Transformations}
Our local coordinates are inspired by generalized  Riemannian (normal) local coordinates \citep{glasmachers2010exponential,lin2023simplifying} and  Fermi coordinates \citep{manasse1963fermi}.
The main idea is to construct local coordinates that handle constraints and facilitate the analytical metric inversion needed for the simplification. 
Using a local coordinate transformation can \emph{iteratively} diagonalize the metric at a \emph{single} evaluation. 
Given a global coordinate, such as a spectral coordinate, we construct, use, and discard a local coordinate system and its coordinate transformation map to the global coordinate at every iteration.
In our approach, the coordinate and its transformation map satisfy these conditions: (i) The map is differentiable 
and satisfies the constraints.
(ii) The local coordinate has no coordinate constraint for
performing RGD  and its origin represents a current iterate in the (global) spectral coordinate.
(iii) The metric evaluated at the origin is an easy-to-inverse matrix, such as an identity \citep{lin2023simplifying} or a diagonal matrix \citep{glasmachers2010exponential}.

{\bf Our Technical Contributions}\,
Existing local coordinates neither account for spectral constraints nor simplify the inverse of the \emph{singular} metric for spectral parameterizations.
We propose new local coordinates for spectral parameterization and demonstrate how they enable spectral-factorized adaptive schemes by satisfying spectral constraints and simplifying the inverse metric computation.
Moreover, we demonstrate that our (nonlinear) scheme preserves the reparametrization invariance (see Claim~\ref{claim:eigen_full_mat_case}).
We also extend our schemes for Kronecker cases.

{\bf Versatility}\,
While we focus on NN training using the GOP, our schemes are suitable for other curvature and applications mentioned in Sec.~\ref{sec:rgd_approach}. For example, our schemes (see  Sec.~\ref{sec:expriments}) can solve gradient-free optimization \citep{wierstra2008natural} and SPD matrix optimization \citep{absil2009optimization}.

\vspace{-0.05cm}
{\bf Notations}
To handle the orthogonal constraint in $\vB$, we use this Cayley map \scalebox{0.8}{$\mathrm{Cayley}(\vN):=(\vI+\vN)(\vI-\vN)^{-1}$}.
Claim~\ref{claim:cayley_map} in the appendix shows why the inversion in the Cayley map always exists.
We introduce a map denoted by \scalebox{0.8}{ $\mathrm{Skew}(\vM):=\vM-\vM^\top$} to make its input $\vM$ skew-symmetric as required by the Cayley map and a lower-triangular restriction denoted by \scalebox{0.8}{$\mathrm{Tril}(\vC)$} to make its input lower-triangular.

\begin{figure*}[!t]
\center
\begin{minipage}[t]{.5\linewidth}
\fbox{	
\begin{minipage}[t]{0.9\linewidth}
		\textbf{Full-matrix  \scalebox{0.8}{   ($\vS = \vB\mathrm{Diag}(\vd)\vB^T$)} }

        \begin{algorithmic}[1]
        \STATE
        \footnotesize  Compute gradient $\vg\coloneqq\nabla \ell(\vmu)$  \\
             \scalebox{0.8}{
             $
            {\vd}  \leftarrow \vd \odot \exp\{ \stepsize_2 \vd^{-1} \odot [ -\gamma \vd +    \mathrm{diag}(\vB^T  \vg\vg^T \vB )  ]  \}
             $
             } \\
             \scalebox{0.72}{   $
             \vB \leftarrow \vB\mathrm{Cayley}(\frac{\stepsize_2}{2}( \mathrm{Skew}(\mathrm{Tril}(\vU))) )
             $
             }
            \STATE
          \scalebox{0.8}{    $
          \vmu \leftarrow \vmu  - \stepsize_1 \vB \mathrm{Diag}(\vd^{-1/p})\vB^T \vg
 $}
				\end{algorithmic}
            \end{minipage}
}

\vspace{0.25cm}
\fbox{
\begin{minipage}[t]{0.9\linewidth}
\textbf{Kronecker (\scalebox{0.6}{$\vS = {\alpha } [\vB^{(C)} \mathrm{Diag}(\vd^{(C)}) (\vB^{(C)})^T] \otimes [\vB^{(K)} \mathrm{Diag}(\vd^{(K)}) (\vB^{(K)})^T]$
   })}	
   \vspace{-0.4cm}
   \begin{algorithmic}[1]
            \STATE
            \footnotesize  Compute gradient \scalebox{0.8}{$\vG\coloneqq \mathrm{Mat}(\nabla \ell(\vmu)) \in \mathcal{R}^{n \times m}$ } \\
             \scalebox{0.78}{   $     \vm^{(l)} = (\vd^{(l)})^{-1} \odot [ - \gamma\vd^{(l)} + \frac{1}{ {\alpha} k^{(l)}}   \mathrm{diag}(\vW^{(l)} )  ]$ } \\
         \scalebox{0.78}{   $ \vd^{(l)}  \leftarrow \vd^{(l)} \odot \exp\{
             \stepsize_2 [\vm^{(l)} - \mathrm{mean}(\vm^{(l)})]
             \} $} \\
             \scalebox{0.8}{ $ \vB^{(l)} \leftarrow \vB^{(l)} \mathrm{Cayley}(\frac{\stepsize_2}{ {2\alpha} k^{(l)} }\mathrm{Skew}( \mathrm{Tril}(\vU^{(l)})  ) ) $ } \\
             \scalebox{0.8}{   $
             \alpha \leftarrow \alpha \exp( \frac{\stepsize_2}{2 } [\mathrm{mean}(\vm^{(C)}) + \mathrm{mean}(\vm^{(K)})] )
             $ } \\
            \STATE
          \scalebox{0.8}{    $\mathrm{Mat}(\vmu)
           \leftarrow \mathrm{Mat}(\vmu) - \stepsize_1 \big({ \alpha^{-1/p}} \big)(\vS^{(C)})^{-1/p}\vG (\vS^{(K)})^{-1/p}
 $}
				\end{algorithmic}
        \end{minipage}
}    
\end{minipage}
\begin{minipage}[t]{.45\linewidth}
\vspace{-0.6cm}
 \caption{
Adaptive update schemes for full-matrix and Kronecker-based spectral factorizations. 
{\bf Full-matrix scheme:} \scalebox{0.8}{ $\mathrm{Tril}(\vU)$} is a lower-triangular matrix with the $(i,j)$-th entry \scalebox{0.8}{ $[\vU]_{ij}:= -[\vB^T \vg\vg^T \vB]_{ij}/(d_i - d_j) $ } when $d_i \neq d_j$ and $0$ otherwise. {\bf  Kronecker-based scheme: } 
We assume that NN weights take a matrix form:\scalebox{0.8}{ $\vM:=\mathrm{Mat}(\vmu) \in \real^{n \times m}$}, where \scalebox{0.8}{$\mathrm{Mat}(\cdot)$} is a matrix representation of vector $\vmu$.
We define  \scalebox{0.78}{$\vW^{(K)} :=(\vB^{(K)})^T \vG^T { (\vS^{(C)})^{-1} } \vG \vB^{(K)}$,  $\vW^{(C)} :=(\vB^{(C)})^T \vG { (\vS^{(K)})^{-1} } \vG^T \vB^{(C)}$,} \scalebox{0.8}{
$k^{(K)}:=n$ and 
$k^{(C)}:=m$ }.
  where \scalebox{0.8}{ $\vS^{(l)} := \vB^{(l)} \mathrm{Diag}(\vd^{(l)}) (\vB^{(l)})^T $ } for \scalebox{0.8}{$l \in \{C,K\}$}.
We define a lower-triangular matrix \scalebox{0.8}{$\mathrm{Tril}(\vU^{(l)})$ }  with its $(i,j)$-th entry \scalebox{0.8}{ $[{\vU^{(l)} }]_{ij}:= -[W^{(l)}]_{ij}/(d^{(l)}_i - d^{(l)}_j) $} if \scalebox{0.8}{$d^{(l)}_i\neq d^{(l)}_j $} and $0$ otherwise, where $d^{(l)}_i$ denotes the $i$-th entry of vector $\vd^{(l)}$.  
For numerical stability, we set \scalebox{0.8}{$[\vU^{(l)}]_{ij}=0$} if \scalebox{0.8}{$|d^{(l)}_i - d^{(l)}_j|$} is near $0$.
See Fig~\ref{fig:kron_updates} in Appx.~\ref{app:connections} for a simplified version.
}\label{fig:kronecker}
\end{minipage}
\vspace{-0.5cm}
\end{figure*}

\vspace{-0.3cm}
\subsection{Full-matrix Schemes via Full Gaussian Approx.}
\label{sec:full_mat_update}
\vspace{-0.15cm}
We present our scheme in the context of full-matrix preconditioners and use the GOP as curvature information. While full-matrix methods are generally impractical for modern NNs, this will serve to illustrate the core ideas that we later apply to structured cases.
Our scheme is given in the top box of Fig.~\ref{fig:kronecker}.
It is a spectral-factorized adaptive scheme for solving problem \eqref{eq:org_opt} without matrix decomposition.
Thus, our scheme enables more efficient and stable root computation than the original one in Eq.~\eqref{eq:root_free}.
Surprisingly, our scheme coincides with the original one (up to first order):

\vspace{-0.05cm}
\begin{claim}
\label{claim:eigen_full_mat_case}
 \textbf{Reparametrization Invariance:} 
  Our update scheme for $\vS$ in the top box of Fig~\ref{fig:kronecker} is equivalent to the scheme in Eq.~\eqref{eq:root_free} up to first-order in terms of $\beta_2$ when $\vd$ does not have repeated entries (proof in Appx.~\ref{app:proof_lemma_full}).
\end{claim}

\vspace{-0.2cm}
To obtain the update scheme (top box of Fig.~\ref{fig:kronecker}), we consider the following procedure to convert an RGD step into an adaptive scheme.
Our procedure follows similar steps in Sec.~\ref{sec:rgd_approach} except for using local coordinates.

\vspace{-0.05cm}
{\bf Step 1}\, We consider a Gaussian problem in \eqref{eq:ref_opt} using 
a learnable spectral parametrization, \scalebox{0.8}{
 $\vS=\vB\mathrm{Diag}(\vd)\vB^T$}, of the inverse covariance $\vS$, where \scalebox{0.8}{$\mathcal{Q}_q=-\half \log \mathrm{det}(\vS)=-\half \sum_i (\log d_i)$}.
 
\vspace{-0.4cm}
\resizebox{0.85\linewidth}{!}{
  \begin{minipage}{\linewidth}
\begin{align}
    \min_{\mu, d, B }\, & \mathcal{L} (\vmu,\vd,\vB):= E_{w \sim q(w;\mu,d,B)}[ \ell(\vw) ] - \gamma \mathcal{Q}_{q}, \nonumber \\
    \textrm{s.t.}\, & \vB\vB^T = \vB^T\vB =\vI\, \text{ and } \vd >0 .
    \label{eq:eig_opt}
\end{align}
\end{minipage}
}

{\bf Step 2}\, We construct a new local coordinate at each iteration to remove the (spectral) constraints in Eq.~\eqref{eq:eig_opt}  when performing RGD.
We then take an unconstrained RGD step in this local coordinate, translate the update to the spectral coordinate, and discard the local coordinate. This coordinate generation process shares the same spirit as Cartan's method of moving frames \citep{ivey2003cartan}.

{\bf Step 2.1}\, Concretely, at iteration $k$, we create a local coordinate $\veta:=(\vdelta,\vm,\vM)$ at the current point $\vtau_k:=(\vmu_k,\vd_k, \vB_k)$ and use this local transformation map

 \vspace{-0.3cm}
\resizebox{0.95\linewidth}{!}{
  \begin{minipage}{1.2\linewidth}
\begin{align}
\vtau(\veta;\vtau_k) :=
\begin{bmatrix}
\vd (\vm;\vtau_k)  \\
\vB (\vM;\vtau_k) \\
\vmu (\vdelta;\vtau_k) \\
\end{bmatrix} = \begin{bmatrix}
    \vd_{k} \odot \exp( \vm ) \\
\vB_{k} \mathrm{Cayley}(\mathrm{Skew}(\mathrm{Tril}(\vM))) \\
    \vmu_k + \vB_k \mathrm{Diag}(\vd_k^{-1/2}) \vdelta 
\end{bmatrix},
\label{eq:translation_map}
\end{align}
\end{minipage}
}

\vspace{-0.2cm}
to translate the change from the local coordinate to the spectral coordinate (see  Claim~\ref{claim:eig_constraint2}), where $\vtau_k=(\vmu_k,\vd_k,\vB_k)$ is considered as a constant in this map and $\odot$ denotes the elementwise product. %
Inspired by the Riemannian normal coordinates \citep{lin2023simplifying},
we construct the map \eqref{eq:translation_map} so that Claims 
\ref{claim:eig_constraint2}
and 
\ref{claim:fisher_full} are satisfied.

\begin{claim}
\label{claim:eig_constraint2}
The map in \eqref{eq:translation_map} satisfies the constraints in \eqref{eq:eig_opt}.
\end{claim}
\vspace{-0.1cm}

\begin{comment}
\begin{claim}
\label{claim:eig_constraint}
{\bf Uniqueness of the Local Coordinates}
The map in  \eqref{eq:translation_map}, which is like the eigendecomposition,
is one-to-one if $\vd$ has no repeated entries
(see Appx.~\ref{app:claim4_proof} for a proof).
\end{claim}
\end{comment}

{\bf Step 2.2}\,
We then take an (unconstrained) RGD step in the local coordinate $\veta$,

 \vspace{-0.5cm}
\resizebox{0.9\linewidth}{!}{
  \begin{minipage}{\linewidth}
\begin{align}
    \mathrm{RGD}: \, \veta_{\text{new}}& \leftarrow \veta_{\text{cur}} - \stepsize [\vF_\eta(\veta_\text{cur}) ]^{-1} \nabla_\eta \mathcal{L}\big|_{\eta:=\eta_{\text{cur}}}, \nonumber \\
    & =\mathbf{0} - \stepsize [\vF_\eta(\mathbf{0}) ]^{-1} \nabla_\eta \mathcal{L}\big|_{\eta:=0} ,
    \label{eq:ngd_local_full}
\end{align}
\end{minipage}
}

\vspace{-0.2cm}
and translate the change $\veta_{\text{new}}$
from the local coordinate

 \vspace{-0.45cm}
\resizebox{0.95\linewidth}{!}{
  \begin{minipage}{\linewidth}
\begin{align}
    \vtau_{k+1} \leftarrow \vtau(\veta_{\text{new}};\vtau_k),
    \label{eq:ngd_global_full}
\end{align}
\end{minipage}
}

to the spectral coordinate, where
the Fisher-Rao metric $\vF_\eta(\veta_\text{cur})$ evaluated at $\veta_{\text{cur}}$ is diagonal  (see  Claim~\ref{claim:fisher_full}) in the local coordinate and the origin
$\veta_{\text{cur}} \equiv \mathbf{0}$ represents the current \scalebox{0.9}{ $\vtau_k=\vtau(\veta_{\text{cur}};\vtau_k)\equiv\vtau(\mathbf{0};\vtau_k)$} in the spectral coordinate.
Notatbly, evaluating the metric at the origin can greatly simplify the metric computation.

{\bf Step 3}\, We obtain the scheme in
the top box of Fig.~\ref{fig:kronecker} with $p=1$
by simplifying RGD in \eqref{eq:ngd_local_full}-\eqref{eq:ngd_global_full}, and making the approximations in \eqref{eq:delta_approx}
(see  Appx.~\ref{sec:deri_full}).
Notably, this approximation can be changed to include other types of curvature information while keeping the root computation efficient.
Our approach efficiently computes  other $p$-roots while existing works \citep{lin2021tractable,
lin2023simplifying,
lincan2024,
tran2021variational,
tan2021analytic,godichon2024natural} are limited to the $1$-root.

Our procedure satisfies parameter constraints (see  Claim \ref{claim:eig_constraint2}) and simplifies the metric inversion (see  Claim  \ref{claim:fisher_full}). 
The simplification is easy because the metric
\scalebox{0.8}{
$\mathbf{F}_\eta(\veta_\text{cur}):=\mathbf{F}_\eta(\veta)\big|_{\eta=\eta_\text{cur}}$} evaluated at the origin (in the local coordinate) is diagonal. 
The metric diagonalization allows us to simplify the inverse metric computation in Eq.~\eqref{eq:ngd_local_full}
even when the metric is singular (see Appx.~\ref{app:handling_repated_d} for a discussion).
Moreover, the gradient 
\scalebox{0.8}{
$\nabla_\eta \mathcal{L}\big|_{\eta=\eta_{\text{cur}}} $}
required by RGD is easy to compute via the chain rule and has an analytical expression (see  Appx.~\ref{sec:deri_full}). 
Furthermore, our update scheme supports other types of curvature information arising from applications, such as gradient-free optimization and SPD matrix optimization (see Sec.~\ref{sec:expriments}).

\begin{claim}
\label{claim:fisher_full}
{\bf{Closed-form Metric Diagonalization}}
    The exact Fisher-Rao metric $\vF_\eta(\veta_\text{cur})$ (for a full-matrix Gaussian) evaluated at the origin \scalebox{0.8}{ $\veta_\text{cur}\equiv\mathbf{0}$} is \emph{diagonal} and has a closed-form expression   (see Eq.~\eqref{eq:fim_full_local} in  Appx.~\ref{app:claim_fisher_full}).
\end{claim}

\begin{figure*}
  \centering
  \includegraphics[width=0.92\linewidth]{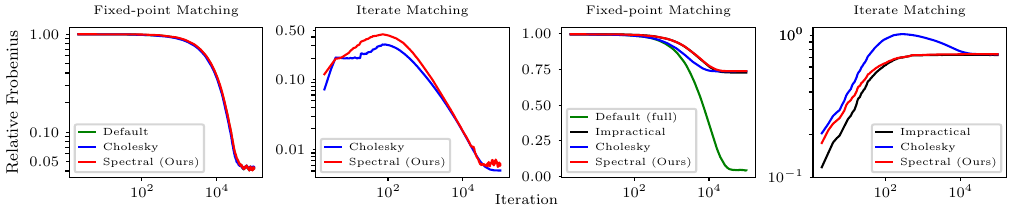}
  \vspace{-3ex}
  \caption{
  Empirical validation of our update schemes for SPD curvature learning.
{\bf Full-matrix Scheme: } 
  The first plot on the left shows that the scheme converges to a fixed-point solution as fast as the default scheme in Eq.~\eqref{eq:root_free} (with $\gamma=1$) to update \scalebox{0.8}{ $\vS\in \real^{100 \times 100}$} and the Cholesky-based scheme. 
  The second plot illustrates how closely our scheme matches the iterates generated by the default update scheme at each iteration.
{\bf Kronecker-based Scheme: } 
  The third plot shows that our update scheme gives a structural approximation \scalebox{0.8}{ $ \vS^{(C)} \otimes \vS^{(K)}$} of a fixed-point solution obtained by the default full-matrix update scheme for \scalebox{0.8}{$\vS \in \mathcal{R}^{99 \times 99}$}, where \scalebox{0.8}{$\vS^{(C)} \in \real^{9 \times 9} $} and \scalebox{0.8}{ $ \vS^{(K)} \in \real^{11 \times 11}$}.
  Our scheme converges as fast as Kronecker-structured baseline methods, including the impractical projection-based method.
  The last plot illustrates how closely our scheme matches the unstructured iterates generated by the default one at each iteration. 
   See Figs.~\ref{fig:full_mat_toy}-\ref{fig:kron_toy} for more results.
  }
 \label{fig:full_merged}
  \vspace{-0.5cm}
\end{figure*}

\begin{comment}
\vspace{-0.2cm}
{\bf  Leveraging the GOP structure}
Since the GOP is semi-definite and $\stepsize_2$ is small, we obtain a linear update for $\vd$:
%
\scalebox{0.85}{
$\vd \odot \exp( \stepsize_2 \vd^{-1} \odot \vh ) \approx  (1-\stepsize_2\gamma)\vd + \stepsize_2 \mathrm{diag}(\vB^T \vg\vg^T \vB)$
} by truncating the exponential map 
\scalebox{0.9}{ $
\vd \odot \exp(\stepsize_2 \vd^{-1} \odot \vh )$} \scalebox{0.9}{$  \approx \vd \odot (1+ \stepsize_2 \vd^{-1} \odot \vh) = \vd + \stepsize_2 \vh
$ }
while keeping $\vd$ positive, where \scalebox{0.9}{ $\vh:=-\gamma \vd + \mathrm{diag}(\vB^T \vg\vg^T \vB)$}.
This makes our approach more similar to the original update in Eq.~\eqref{eq:root_free}. 
\end{comment}

\vspace{-0.35cm}
\subsection{
Connections to Diagonal Adaptive  Methods}
\label{sec:diag_case_rgd}
\vspace{-0.2cm}
Our full-matrix scheme in Fig.~\ref{fig:kronecker} also applies in diagonal cases by forcing $\vB$ to be a diagonal matrix. 
We achieve that by changing map $\mathrm{Tril}(\cdot)$ to $\mathrm{Diag}(\cdot)$ in the update rule. Consequently, $\vB$ becomes a constant identity matrix (up to sign changes) that can be ignored.
Similar to the full matrix case, we can obtain this scheme through a diagonal Gaussian approximation.
When truncating the exponential map, our scheme becomes the 
$1$-root variant of RMSprop and AdaGrad  \citep{lincan2024} for $p=1$.
If applying other fractional $p$-roots, our scheme also recovers RMSprop and AdaGrad for $p=2$ and the fractional diagonal method \citep{chen2021closing} for $p=4$. See Appx.~\ref{app:connection_diag} for the details.

\vspace{-0.35cm}
\subsection{Kronecker Schemes via Matrix Gaussian Approx.}
\label{sec:mat_gauss_nn}
\vspace{-0.2cm}
Using Kronecker-structured preconditioners \citep{martens2015optimizing,li2017preconditioned,
gupta18shampoo} is necessary for large models, as a full-matrix preconditioner is too large to store.
Many Kronecker-based methods \citep{zhang2018noisy,ren2021tensor,lin2023simplifying,lincan2024} are based on a (matrix) Gaussian family with Kronecker-structured inverse covariance \scalebox{0.8}{$\vS=\vS^{(C)} \otimes \vS^{(K)}$}.
Thus, we want to extend the procedure in Sec.~\ref{sec:full_mat_update} to Kronecker cases. 
However, the Fisher-Rao metric for a Kronecker structure is singular
and non-block-diagonal because the Kronecker factorization is not unique.
Thus, we cannot use the procedure in Sec.~\ref{sec:full_mat_update} to obtain an adaptive scheme.
Existing works such as \citet{zhang2018noisy,
lin2019fast,
lincan2024} make an additional approximation for the metric to tackle this.
Instead, we overcome this without approximation by imposing a determinant constraint on each Kronecker factor and introducing a learnable scalar $\alpha$ to make the factorization unique (see  Claim~\ref{claim:kron_unique}) and simplify the inverse metric computation.

\vspace{0.05cm}
\begin{claim}
\label{claim:kron_unique}
    A Kronecker-structured positive-definite matrix $\vS$ can be uniquely expressed as  
    \scalebox{0.8}{ $\vS=\alpha [\vS^{(C)} \otimes \vS^{(K)}]$ } with constraints  \scalebox{0.8}{$\mathrm{det}(\vS^{(C)})=\mathrm{det}(\vS^{(K)})=1$ } and \scalebox{0.8}{$\alpha>0$}. 
    (see Appx.~\ref{app:claim6_proof})
\end{claim}

\vspace{-0.2cm}
We then propose a spectral parametrization and local coordinates for each Kronecker factor 
 \scalebox{0.8}{
$\vS^{(l)}=\vB^{(l)}\mathrm{Diag}(\vd^{(l)})(\vB^{(l)})^T$}
for  \scalebox{0.8}{$l \in \{C,K\}$}, where 
 \scalebox{0.8}{
$\mathrm{det}(\mathrm{Diag}(\vd^{(l)}))=1$} is  the  determinant constraint.
Thanks to our local coordinates, we can again  simplify the inverse metric computation (see  Claim~\ref{claim:fisher_kron}).
We then follow a similar procedure in Sec.~\ref{sec:full_mat_update} to obtain an update scheme for Kronecker cases (see bottom box of Fig.~\ref{fig:kronecker}).

\vspace{-0.05cm}
{\bf Handling New Constraints via \emph{Local} Transformations}
For the positive scalar $\alpha$, we introduce a local coordinate $n$ and use an exponential map in the coordinate transformation: $\alpha(n; \alpha_k) = \alpha_k \exp(n)$ at each iteration $k$.
 We drop the Kronecker factor index $l$ for simplicity. The coordinate transformation map for each factor is similar to the map in full-matrix cases, expect that for vector 
\scalebox{0.9}{
$\vd(\vm; \vd_k
    )  = \vd_{k} \odot \exp( \vm )$}, 
we require  \scalebox{0.9}{$\mathrm{sum}(\vm)=0$} to satisfy the determinant constraint (i.e.,
\scalebox{0.8}{$\mathrm{det}(\mathrm{Diag}(\vd(\vm)))=1$}), where $j$ is the length of vector $\vd_k$ and the local coordinate 
\scalebox{0.8}{
$\vm := [m_1,\dots,m_{j-1}, -\sum_{i}^{j-1}m_i]$} has only \scalebox{0.8}{$(j-1)$} free variables.

\begin{claim}
\label{claim:fisher_kron}
{\bf{Closed-form Metric Block Diagonalization}}
    The Fisher-Rao metric \scalebox{0.8}{ $\vF_\eta(\veta_\text{cur})$} (for a  matrix Gaussian) 
 and its inverse evaluated at \scalebox{0.8}{$\veta_\text{cur}\equiv\mathbf{0}$} are block-diagonal and have closed form (see Eq.~\eqref{eq:fim_kron_local} in Appx.~\ref{app:kron_fim}).
\end{claim}

\begin{comment}
\vspace{-0.25cm}
{\bf  Leveraging the GOP structure}
Similar to the full-matrix case, we obtain the following update by leveraging the GOP structure while satisfying the constraints in $\vd^{(l)}$ and $\alpha$.

\vspace{-0.45cm}
\resizebox{0.92\linewidth}{!}{
  \begin{minipage}{\linewidth}
\begin{align*}
    \vm^{(l)} & = (1-\stepsize_2\gamma ) \vd^{(l)} +  \frac{\stepsize_2}{ {\alpha} k^{(l)}}  \mathrm{diag}(\vW^{(l)} ) \\
   \vd^{(l)} & \leftarrow \exp( \log( \vm^{(l)} ) -\mathrm{mean}(\log( \vm^{(l)} )) )\\
  \alpha & \leftarrow \exp\big(  \mathrm{mean}(\log( \vm^{(C)} ))/2 + \mathrm{mean}(\log( \vm^{(K)} ))/2
  \big) 
\end{align*}
\end{minipage}
}

where  \scalebox{0.85}{$k^{(l)}$} and  \scalebox{0.85}{$\vW^{(l)}$} are defined at the caption of Fig.~\ref{fig:kronecker} and \scalebox{0.85}{$l \in \{C,K\}$}.
See Appx.xxx for more details.  \todo{add the appendix}
\end{comment}

%
%

\begin{figure*}
  \centering
  \includegraphics[width=\linewidth]{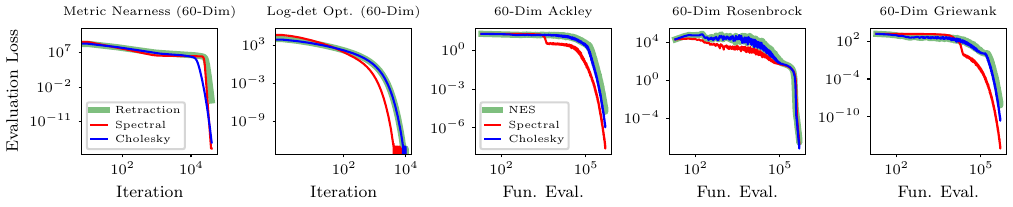}
  \vspace{-6ex}
  \caption{
  Experiments showcase the efficacy and versatility of our approach for generic curvature learning.
   Our update scheme matches the equivalent Riemannian baselines, empirically illustrating the reparametrization invariance.
  {\bf SPD Matrix Optimization:}
  The first two plots on the left show the performance of our full-matrix update scheme for learning SPD matrices. Our update scheme matches the baselines, as our scheme is RGD in local coordinates.
  {\bf Gradient-free Optimization:}
  The last three plots show the performance of our scheme for gradient-free optimization problems on Ackley (multimode), Rosenbrok (flat valley), and Griewank (multimode) functions.
  See Fig.~\ref{fig:extra_results} in the appendix for more results.
  }
  \label{fig:other_res}
  \vspace{-0.5cm}
\end{figure*}

\vspace{-0.35cm}
\subsection{Practical Considerations for NN Training}
\label{sec:practical_NN}
\vspace{-0.2cm}
{\bf Using a Truncated Cayley Map
for Low-precision Training
}
Recall that our schemes in Fig.~\ref{fig:kronecker} use the Cayley map that involves matrix inversion.
To  work with half-precision and further reduce the computational cost, 
we use a truncated Cayley map, similar to \citet{  liu2021orthogonal,li2020efficient,qiu2023controlling}, 
for NN problems. 
Our truncation is based on a Neumann series for the matrix inversion \citep{krishnan2017neumann,lorraine2020optimizing,qiu2023controlling}. This is possible because we can approximate the matrix inversion in the Cayley map
\scalebox{0.8}{
 $ \mathrm{Cayley}(\stepsize \vN) = (\vI+\stepsize \vN) (\vI-\stepsize\vN)^{-1}
 = (\vI+\stepsize \vN) \prod_{l=0}^{\infty} (\vI+ (\stepsize\vN)^{2^l}) $} \scalebox{0.8}{ $ \approx (\vI+\stepsize \vN)^2 (\vI+ (\stepsize \vN)^2)(\vI+ (\stepsize \vN)^4)
 $}
based on a convergent Neumann series,  when $\stepsize$ is small enough so that the Frobenius norm $\|\stepsize \vN\|_{\text{Frob}}<1$.
Consequently, we use a nonconstant step size, similar to \citet{lincan2024}, to keep the norm bounded as required by the truncation.
Fig.~\ref{fig:exact_vs_truncated} in the appendix shows that using the truncation significantly reduces running time while maintaining performance.

\vspace{-0.4cm}
\paragraph{
Stabilizing Training using Preconditioned Gradient Clipping or Grafting}
For training transformers, it is common practice to use preconditioned gradient clipping \citep{liu2023sophia,shen2024variational} or grafting \citep{agarwal2019efficient,shi2023distributed,vyas2024soap} so that
the norm of their descent direction is either bounded layerwisely \citep{zhang2024transformers} or bounded as the norm of Adam \citep{shi2023distributed,vyas2024soap}.
This is needed because of heavy-tailed noise \citep{zhang2020adaptive}. We use the preconditioned gradient clipping in our experiments to train transformers.
Other approaches like sign descent \citep{chen2023symbolic}, normalized gradient descent \citep{cutkosky2020momentum}, and 
spectral norm descent
\citep{bernstein2024old,muon2024}
also implicitly bound the norm of their descent direction.

\vspace{-0.4cm}
\section{Experiments}
\label{sec:expriments}

\vspace{-0.2cm}
\subsection{SPD Curvature Learning for the GOP}
\vspace{-0.15cm}
{\bf Empirical Validation of the Full-matrix Update Scheme}
First, we empirically evaluate our scheme for
using the GOP as curvature information to demonstrate how our update scheme relates to the default scheme (see Eq.~\eqref{eq:root_free}) by indirectly learning
\scalebox{0.8}{$\vS^{\text{(spectral)}}:=\vB\mathrm{Diag}(\vd)\vB^T$}.
We compare our scheme to the default one on $\vS$ as
\scalebox{0.8}{$ \vS_{k+1}^{\text{(default)}} \leftarrow (1-\stepsize)  \vS_{k}^{\text{(default)}} + \stepsize \vg_k\vg_k^T$}, and the inverse-free scheme \citep{lincan2024} for a learnable Cholesky factor \scalebox{0.8}{$\vC$} of \scalebox{0.8}{$\vS^{-1}$} (i.e., \scalebox{0.8}{ $\vS^{\text{(cholesky)}}:=(\vC\vC^T)^{-1}$}).
We focus on the curvature estimation of $\vS$ based on a fixed gradient sequence $\{\vg_1,\dots,\vg_T\}$ and initialized by the same $\vS_0$. 
We consider two scenarios:
(1) fixed-point matching and (2) iterate matching. 

\vspace{-0.3cm}
\begin{description}
 \item[Fixed-point matching]\,
 The ground truth  is a fixed-point solution, \scalebox{0.8}{ $\vS_{*} = E[\vg\vg^T]=\vSigma$}, to the default update scheme as \scalebox{0.8}{ $\vS_{*}=(1-\stepsize)\vS_{*} + \stepsize \vg_k\vg_k^T $}, where
  $\vg_k$ is independently generated from a Gaussian distribution \scalebox{0.8}{$\vg_k \sim \gauss(\mathbf{0}, \vSigma)$} at each iteration $k$.
We evaluate each scheme at iteration $k$ by comparing its current estimate denoted by \scalebox{0.8}{$\vS_k^{\text(est)}$} to the fixed point.
We use a relative Frobenius norm 
\scalebox{0.9}{
$\frac{ \|\vS_* - \vS_k^{\text{(est)}}\|_\text{Frob}}{\|\vS_*\|_\text{Frob}}$ }
to measure the difference.

\vspace{-0.2cm}
\item[\bf Iterate matching]
The ground truth is a sequence of matrices \scalebox{0.8}{ $\{\vS_1^{\text{(true)}},\dots,\vS_T^{\text{(true)}}\}$}
generated by the default scheme when applying the scheme to the gradient sequence.
We want to match the iterate the default scheme generates at every step. 
We use a relative Frobenius norm 
\scalebox{0.9}{
$\frac{ \|\vS_k^{\text{(true)}} - \vS_k^{\text{(est)}}\|_\text{Frob}}{\|\vS_k^{\text{(true)}}\|_\text{Frob}}$} 
to measure the discrepancy between an update scheme and the default one at every iteration $k$.
\end{description}

\vspace{-0.4cm}
From the first two plots in  Fig.~\ref{fig:full_merged}, we can see that our update scheme performs similarly to the default update scheme in the two scenarios. Together with Claim~\ref{claim:eigen_full_mat_case}, these results show that our scheme
coincides with the default one theoretically and empirically.
See Appx.~\ref{app:extra_empirical_valida} and Fig.~\ref{fig:full_mat_toy} for more details.

\vspace{-0.2cm}
\paragraph{Empirical Evaluation of the Kronecker-based Update Scheme}
We compare our Kronecker scheme to the default full-matrix scheme on $\vS$: 
$ \vS_{k+1} \leftarrow (1-\stepsize)  \vS_{k} + \stepsize \vg_k\vg_k^T$. %
As baselines, we consider the curvature estimation used in the structured Cholesky factorization \citep{lincan2024}, and an impractical projection-based method~\citep{van1993approximation}: \scalebox{0.8}{ $(\vS_{k+1}^{(C)}, \vS_{k+1}^{(K)}) \leftarrow \mathrm{Proj}( (1-\beta)( \vS_k^{(C)} \otimes \vS_k^{(K)}) + \beta \vg_k\vg_k^T ) $}.
We use a similar experimental setup and consider two similar scenarios
discussed in the full-matrix case.
Here, we initialized all update schemes by a Kronecker structured matrix $\vS_0$ to remove the difference introduced by initialization.

From the last two plots in  Fig.~\ref{fig:full_merged}, we can see that our structural scheme performs as well as existing structural baselines. Our approach performs similarly to the impractical method that requires storing a full matrix and solving a projection optimization problem at every iteration. This illustrates the effectiveness of our approach in Kronecker cases.
See Appx.~\ref{app:extra_empirical_valida} and Fig.~\ref{fig:kron_toy} for more details.

\begin{figure*}
  \centering
  \includegraphics[width=1.0\linewidth]{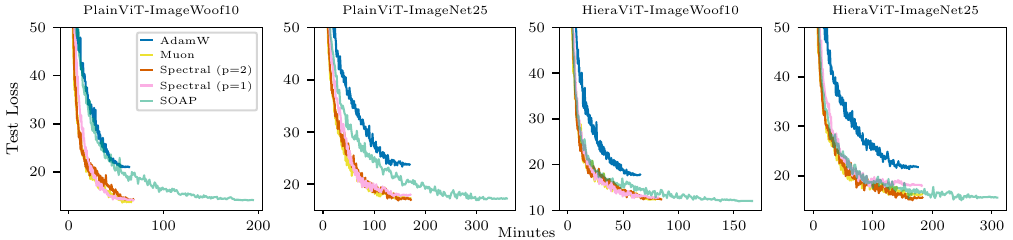}
  \vspace{-5ex}
  \caption{
  Experiments demonstrate the efficiency of our update schemes for low-precision NN training. 
  The plots show the performance of our Kronecker-based scheme for training vision transformers with half precision. 
  All models are trained for 210 epochs, including 10 epochs for warmup.
  For SOAP and our method, we update
  their preconditioners every two iterations.
  SOAP performs much slower than the other methods because it has to run in single precision to use matrix decomposition. 
  Using a different matrix root can affect the performance.
  Our method not only matches Muon's performance but also
  opens the door to using curvature information and matrix roots beyond Muon.
  See  Figs.~\ref{fig:imagewoof} and \ref{fig:imagenet25} in Appx.~\ref{app:extra_nn_training} for a comparison of the methods based on iteration efficiency and wall-clock time. 
 \vspace{-0.4cm} 
  }
  \label{fig:results}
\end{figure*}

\vspace{-0.2cm}
\subsection{Full-matrix Case: SPD Matrix Optimization}
\vspace{-0.15cm}
We consider SPD matrix optimization problems to demonstrate the versatility of our scheme.
We aim to learn an SPD matrix $\vS$ from another type of curvature information that can be negative-definite. Therefore, the (linear) original scheme of $\vS$ in  Eq.~\eqref{eq:root_free}  is unsuitable in this setting because the original scheme assumes the curvature (i.e., the GOP) is positive-semi-definite.
In Riemannian optimization, we introduce a non-linear retraction map
in the original scheme so that the updated $\vS$ is guaranteed to be SPD. This is known as RGD with retraction \citep{absil2009optimization,boumal2014manopt}.
We consider this retraction-based RGD \citep{boumal2014manopt} and the Cholesky-based RGD \citep{lin2023simplifying} as baselines. We consider a metric nearness problem \citep{brickell2008metric}\scalebox{0.8}{ $\min_{S \succ 0} \ell(\vS):= \frac{1}{2N}\sum_{i=1}^{N} \|\vS\vQ\vx_i-\vx_i\|_2^2$} 
and a log-det optimization problem \citep{han2021riemannian} \scalebox{0.8}{ $\min_{S \succ 0} \ell(\vS):= \mathrm{Tr}(\vS\vQ) - \log \mathrm{det}(\vS)$}
, where $\vQ \in \real^{d \times d}$ is a known SPD matrix and only a subset of $\vx_i \in \real^d$ are observed at each iteration. The optimal solution is $\vS_*=\vQ^{-1}$ for these two problems. We measure the difference between an estimate $\vS_\text{est}$ and the ground truth $\vS_*$ using \scalebox{0.8}{ $\ell(\vS_\text{est})-\ell(\vS_*)$} as the evaluation loss.
We consider a case for $d=60$ and generate $\vQ$ and $\vx_i$.
As we can see from the first two plots in Fig.~\ref{fig:other_res}, our method performs as well as the baseline methods. Our approach works well for estimating SPD matrices even when the curvature information is not SPD. This shows that our approach can  incorporate other types of curvature information for learning $\vS$.

\vspace{-0.3cm}
\subsection{Full-matrix Case: Gradient-free Optimization}
\vspace{-0.15cm}
We consider classic gradient-free problems to demonstrate the flexibility of our approach for preconditioning (i.e., learning $\vmu$ and $\vS$).
We learn $\vS$ from a new type of curvature information to solve an original problem $\min_{\mu}\ell(\vmu)$, where we only allow to
evaluate the value of $\ell(\vmu)$.  We consider natural evolution strategies (NES) \citep{wierstra2008natural} and the Cholesky-based NES \citep{glasmachers2010exponential,fukushima2011proposal,lin2021tractable} as our baselines. Like these methods, we consider a Gaussian problem in Eq.~\eqref{eq:ref_opt} with $\gamma=0$ to approximate the original problem.
We can solve this Gaussian problem using the procedure in Sec.~\ref{sec:full_mat_update},
where we use the $1$-root and the REINFORCE estimator in Eq.~\eqref{eq:delta_approx}, which evaluates the function value of $\ell(\vw)$ and uses samples $\vw$ generated from the Gaussian $q$. We use the techniques of  \citet{wierstra2008natural} and \citet{fukushima2011proposal} to reduce the number of Monte Carlo samples and function evaluations.
We use $\ell(\vmu)$ as the evaluation loss.
From the last three plots of Fig.~\ref{fig:other_res}, we can observe that our update scheme is competitive with existing Riemannian methods. This further illustrates the efficacy and versatility of our approach. See Appx.~\ref{app:bb_opt} and Fig.~\ref{fig:extra_results} for more details.

\vspace{-0.2cm}
\subsection{Kronecker-based Case: Low-precision NN Training}
\vspace{-0.2cm}
Now, we examine our Kronecker-based update scheme for training transformers in half-precision (BFP16). We use our update scheme to train vision transformers with half-precision from scratch. 
Training transformers in half-precision allows us to
evaluate the numerical stability of our approach because matrix inversion and decomposition are unstable and thus unavailable in half precision.  
We then show the effectiveness and efficiency of our approach by comparing our method to strong baselines like AdamW and two recently proposed Shampoo variants, SOAP \citep{vyas2024soap} and Muon \citep{muon2024}.
Both SOAP and Muon are limited to a specific matrix root and Shampoo-based curvature information. In contrast, our approach supports arbitrary matrix roots and generic types of curvature information.
We consider training two vision transformers, PlainViT (85.6M parameters) \citep{beyer2022better} and HieraViT (50.7M parameters) \citep{ryali2023hiera}, on ImageWoof-10 and ImageNet-25 datasets. 
These ViT models only contain transformer modules except for their embedding layers.
All training methods except for SOAP support half-precision training. SOAP has to use single precision (FP32) data types due to matrix decomposition.
To demonstrate the efficiency of our approach, we update the preconditioners of our method and SOAP every two iterations.
We use random search \citep{choi2019empirical} to tune all available hyperparameters for each method on each dataset using 80 runs. See Appx.~\ref{app:extra_nn_training} for the details of the experimental setup.
From the plots in  Fig.~\ref{fig:results},  we can see that our method effectively trains transformers in half-precision and matches the performance of state-of-the-art methods like Muon.  
Compared to SOAP, we can clearly see the computational advantages of
fast root computation without matrix decomposition.
The plots also show that using a different root can affect the performance of matrix methods.
Finally, these results highlight the benefits of non-diagonal methods and underscore the need for further development.
See  Figs.~\ref{fig:imagewoof} and \ref{fig:imagenet25}
 for more results.

\vspace{-0.35cm}
\section{Conclusion}
\vspace{-0.2cm}
We present spectral-factorized curvature learning schemes for designing training methods. Our approach addresses the instability and inefficiency of using matrix fractional roots in low-precision training, enabling non-diagonal methods to support other fractional roots and generic curvature information.
An interesting direction for future work is to evaluate our schemes in large-scale settings and explore the potential benefits of using different  roots and curvature information.
\newpage

\bibliography{refs}
\bibliographystyle{icml2024}

\newpage
\appendix
\onecolumn
\begin{figure*}[!t]
\center
\begin{minipage}[t]{.45\linewidth}
\textbf{Full-matrix (Original)}
\begin{algorithmic}[1]
    \STATE
    \footnotesize  Compute gradient $\vg\coloneqq\nabla \ell(\vmu)$  \\
     \scalebox{0.8}{
     $
    {\vS}  \leftarrow (1-\gamma \stepsize_2)\vS + \stepsize_2  (\vg\vg^T+ \lambda \vI )
     $
     } \\
     \scalebox{0.72}{   $
     \color{blue}\vB \mathrm{Diag}(\vd) \vB^T = \mathrm{Eigen}(\vS)$
     }
    \STATE
  \scalebox{0.8}{    $
  \vmu \leftarrow \vmu  - \stepsize_1 \vB \mathrm{Diag}(\vd^{-1/p})\vB^T \vg
$}
        \end{algorithmic}
\end{minipage}
    \begin{minipage}[t]{.48\linewidth}
\textbf{Ours \scalebox{0.8}{   ($\vS = \vB\mathrm{Diag}(\vd)\vB^T$)} }
\begin{algorithmic}[1]
    \STATE
    \footnotesize  Compute gradient $\vg\coloneqq\nabla \ell(\vmu)$  \\
     \scalebox{0.8}{
     $ {\vd}  \leftarrow (1-\gamma\stepsize_2)\vd + \stepsize_2 \big( \mathrm{diag}(\vB^T  \vg\vg^T  \vB ) + \lambda \big)  $
     } \\
     \scalebox{0.72}{   $ 
     \vB \leftarrow \vB\mathrm{Cayley}(\frac{\stepsize_2}{2}( \mathrm{Skew}({\mathrm{Tril}}(\vU))) )
     $
     }
    \STATE
  \scalebox{0.8}{    $
  \vmu \leftarrow \vmu  - \stepsize_1 \vB \mathrm{Diag}(\vd^{-1/p})\vB^T \vg
$}
        \end{algorithmic}
\end{minipage}
 \vspace{-0.1cm}   \caption{
Comparison between the original full-matrix update scheme (e.g., full-matrix RMSprop when $\gamma=1$) and our update scheme when the exponential map is truncated.  We can see that our update scheme is a decomposition-free version of the adaptive method. Here, we can use the first-order truncation of the exponential map (see the top box of  Fig.~\ref{fig:kronecker} for the map).
This is possible because we leverage the positive semi-definiteness of the GOP. Thanks to the 
positive semi-definitness, the update of $\vd$ is always non-negative.
In practice, we introduce a damping term $\lambda$ so that $\vd$ is always positive.
}\label{fig:full_updates}
\end{figure*}

\begin{figure*}[!t]
\center
\begin{minipage}[t]{.49\linewidth}
\textbf{ 
Kronecker-based 
\citep{lincan2024}} 
        \begin{algorithmic}[1]
    \STATE
    \footnotesize  Compute gradient $\vG\coloneqq \mathrm{Mat}(\nabla \ell(\vmu))$  \\
     \scalebox{0.8}{ $             \vS^{(l)} = (1-\gamma\stepsize_2){\vS}^{(l)}  +\frac{\stepsize_2}{k^{(l)}} \big(\vQ^{(l)} + \Lambda^{(l)}\vI\big)$ } \\
     \scalebox{0.8}{  \color{blue} $ \vB^{(l)} \mathrm{Diag}(\vd^{(l)}) (\vB^{(l)})^T = \mathrm{Eigen}(\vS^{(l)}) $ }  \\
     \scalebox{0.8}{
    $
    \big( \vS^{(l)}\big)^{-1/p} = \vB^{(l)} \mathrm{Diag}\big(( \vd^{(l)})^{-1/p} \big)  (\vB^{(l)})^T
    $ 
     }\\
    \STATE
  \scalebox{0.8}{    $
  \mathrm{Mat}(\vmu) \leftarrow \mathrm{Mat}(\vmu) - \stepsize_1 ({\vS}^{(C)})^{-1/p}\vG ({\vS}^{(K)})^{-1/p}
$}
        \end{algorithmic}
\end{minipage}
    \begin{minipage}[t]{.48\linewidth}
\textbf{Ours (\scalebox{0.6}{ $\vS = {\alpha } [\vB^{(C)} \mathrm{Diag}(\vd^{(C)}) (\vB^{(C)})^T] \otimes [\vB^{(K)} \mathrm{Diag}(\vd^{(K)}) (\vB^{(K)})^T]$
})}

\begin{algorithmic}[1]
    \STATE
    \footnotesize  Compute gradient $\vG\coloneqq \mathrm{Mat}(\nabla \ell(\vmu))$  \\
     \scalebox{0.8}{ $             \vn^{(l)} = (1-\gamma\stepsize_2)\vd^{(l)} + \frac{\stepsize_2}{ {\alpha} k^{(l)}}\big(   \mathrm{diag}((\vB^{(l)})^T \vQ^{(l)} \vB^{(l)}) + \Lambda^{(l)}\big)$ } \\
     \scalebox{0.8}{   $ \vB^{(l)} \leftarrow \vB^{(l)} \mathrm{Cayley}(\frac{\stepsize_2}{2 {\alpha} k^{(l)} }\mathrm{Skew}( \mathrm{Tril}(\vU^{(l)})  ) ) $ } \\
     \scalebox{0.8}{   $ \vd^{(l)}  \leftarrow  \exp\big[\log(\vn^{(l)}) {\color{red} - \mathrm{mean}(\log(\vn^{(l)}))
     }\big] $} \\
     \scalebox{0.8}{   $\color{red}     \alpha \leftarrow \alpha  \exp\big[ \mathrm{mean}(\log(\vn^{(C)}))/2 + \mathrm{mean}(\log(\vn^{(K)}))/2\big]    $ } \\
    \STATE
  \scalebox{0.8}{    $
  \mathrm{Mat}(\vmu) \leftarrow \mathrm{Mat}(\vmu)  - \stepsize_1 \big({ \alpha^{-1/p}} \big)({\vS}^{(C)})^{-1/p}\vG ({\vS}^{(K)})^{-1/p}
$}
        \end{algorithmic}
\end{minipage}
 \vspace{-0.1cm} \caption{
 Comparison between the Kronecker-based update scheme \citep{lincan2024} with $p=1$ and our update scheme when the exponential map is truncated.  
 We define  
\scalebox{0.8}{ $\vQ^{(C)}:=\vG ({\vS}^{(K)})^{-1}\vG^T$,
 $\vQ^{(K)}:=\vG^T ({\vS}^{(C)})^{-1}\vG$}, and \scalebox{0.8}{ $\vW^{(l)}=(\vB^{(l)})^T \vQ^{(l)} \vB^{(l)}$}, 
 where 
 \scalebox{0.8}{
$\vS^{(l)}:=\vB^{(l)}\mathrm{Diag}(\vd^{(l)}) \big(\vB^{(l)}\big)^T$}.
 Unlike \citet{lincan2024}, we consider a unique representation of preconditioner $\vS$ by introducing a scalar $\alpha$ and a determinant constraint on $\vd^{(l)}$.
 The lines highlighted in red are used to ensure the constraints.
 We can use the first-order truncation of the exponential map (see the bottom box of  Fig.~\ref{fig:kronecker} for the map) because we exploit the positive semi-definiteness of the curvature information $\vQ$ used in \citet{lincan2024}.
 Thus, $\vn^{(l)}$ is non-negative even when using the truncation.
 To make $\vn^{(l)}$ positive,  we use an adaptive damping term, \scalebox{0.8}{
  $\Lambda^{(l)} := \frac{\lambda \mathrm{Tr}\big((\vS^{(C)})^{-1}\big)\mathrm{Tr}\big((\vS^{(K)})^{-1}\big)}{\mathrm{Tr}\big((\vS^{(l)})^{-1}\big)} =
  \frac{\lambda k^{(l)} \mathrm{mean}\big((\vd^{(C)})^{-1}\big)\mathrm{mean}\big((\vd^{(K)})^{-1}\big)}{\mathrm{mean}\big((\vd^{(l)})^{-1}\big)}
 $}, as suggested by \citet{lincan2024}.
We can even use a truncated Cayley map discussed in Sec.~\ref{sec:practical_NN} for low-precision training. In this case, we update $\vn^{(l)}$ using a common $\stepsize_2$ while updating $\vB^{(l)}$ using a block-specific  
nonconstant $\stepsize_2^{(l)}:=\frac{ \bar{ \stepsize}_2 \alpha \kappa^{(l)} }{ \| 
  \mathrm{Skew}( \mathrm{Tril}(\vU^{(l)})  ) 
 \|_{\text{Frob}} }$ required by the truncation, where scalar $\bar{
\stepsize_2}$ remains constant for all blocks $\vB^{(l)}$.
Similar to \citet{vyas2024soap} and \citet{lincan2024}, our update scheme only introduces one additional scalar hyperparameter and supports momentum and weight decay.
}\label{fig:kron_updates}
\end{figure*}

\section{
Leveraging 
Positive Semi-definiteness 
for Simplification via Truncating the Exponential Map
}
\label{app:connections}

Our update scheme supports generic curvature information to estimate a spectral-factored SPD matrix $\vS$. When the curvature information is 
positive semi-definite, we can further simplify our scheme by using a truncated exponential map.
We consider the following cases for GOP-based curvature information.

\subsection{Diagonal Case}
\label{app:connection_diag}
Here, we connect our scheme with  $\gamma=1$ to the RMSprop method.
Similarly, we can connect our scheme to AdaGrad by setting $\gamma=0$.
Observe that eigenvalues are diagonal entries of a diagonal preconditioning matrix (i.e., $\vd=\mathrm{diag}(\vS)$).
Because $\vB$ is now a diagonal and orthogonal matrix, its diagonal entries can only be 1 or -1.
Using this result, we can further simplify our update scheme with $p=2$ and recover the RMSprop update rule when using a first-order truncation of the exponential map.
\begin{align}
  \vd & \leftarrow  \vd \odot \exp\{ \stepsize_2\, \vd^{-1} \odot [ -\gamma \vd +    \mathrm{diag}(\vB^T  \vg\vg^T \vB )  ]  \} \approx \vd + \stepsize_2 [ -\gamma \vd +    \mathrm{diag}( \vB^T \vg\vg^T \vB )  ]   = (1-\stepsize_2\gamma) \vd + \stepsize_2 \mathrm{diag}( \vg\vg^T )   \nonumber \\
  \vB & \leftarrow \vB\mathrm{Cayley}(\frac{\stepsize_2}{2}\, \mathrm{Skew}({ \color{red}\mathrm{Diag}(\vU)} ) ) = \vB \mathrm{Cayley}(\mathbf{0}) =\vB \nonumber \\
  \vmu &\leftarrow \vmu  - \stepsize_1\, \vB \mathrm{Diag}(\vd^{-\nicefrac{1}{p}})\vB^T \vg =
         \vmu  - \stepsize_1\, \mathrm{Diag}(\vd^{-\nicefrac{1}{p}}) \vg
         \label{eq:root_free_diag},
\end{align} where $\mathrm{diag}(\vB^T  \vg\vg^T \vB )=\mathrm{diag}(  \vg\vg^T )$, $\vB \mathrm{Diag}(\vd^{-\nicefrac{1}{p}})\vB^T=\mathrm{Diag}(\vd^{-\nicefrac{1}{p}})$, and we use a first-order truncation of the exponential map $\vd \odot \exp(\vd^{-1} \odot \vn) \approx \vd \odot (\mathbf{1} + \vd^{-1} \odot \vn)= \vd + \vn$.
Due to the skew-symmetrization, $\mathrm{Skew}({ \mathrm{Diag}(\vU)} )$ is always a zero matrix. Thus, according to our update scheme, $\vB$ remains unchanged in diagonal cases because $\mathrm{Cayley}(\mathbf{0})=\vI$.

\subsection{Full-matrix Case}
We consider a truncated exponential map to bring our update schemes closer to the original ones.
Similar to the diagonal case, we use the first-order truncation of the exponential  map and obtain the update in
the rightmost box of Fig.~\ref{fig:full_updates}.
From Fig.~\ref{fig:full_updates}, we can observe the similarity between the original scheme  and ours.

\subsection{Kronecker-based Case}
We can further simplify our update scheme in Fig.~\ref{fig:kronecker} when truncating the exponential map.
To see that, we give Claims \ref{claim:approx1} and \ref{claim:approx2}.
According to Claim~\ref{claim:approx2},  our update scheme in Fig.~\ref{fig:kronecker} can be merged and reexpressed as
\begin{align*}
\alpha^2 \mathrm{Diag}(\vd^{(C)}) \otimes \mathrm{Diag}(\vd^{(K)}) &= \mathrm{Diag}(\alpha \vd^{(C)}) \otimes \mathrm{Diag}(\alpha \vd^{(K)}) \\
& \approx  ( \alpha \exp \big[\mathrm{mean}( \log(\vn^{(C)}) )/2 + \mathrm{mean}( \log(\vn^{(K)}) )/2 \big]  )^2 
\mathrm{Diag}( \exp \big(\vv^{(C)}\big)) \otimes 
\mathrm{Diag}(
\exp\big(\vv^{(K)}\big)),
\end{align*} where $\vv^{(l)}:=\log(\vn^{(l)}) - \mathrm{mean}( \log(\vn^{(l)}) )$ and
$\vn^{(l)}:=
(1-\gamma \stepsize_2)\vd^{(l)} +  \frac{\stepsize_2}{\alpha k^{(l)}} \mathrm{diag}\big(\vW^{(l)} \big)
$.
We then split the above   update into each individual factor  as 
\begin{align*}
   \vd^{(l)} & \leftarrow \exp(\vv^{(l)}) = \exp( \log(\vn^{(l)}) - \mathrm{mean}(\log(\vn^{(l)})) ) \\
   \alpha & \leftarrow \alpha \exp(
   \mathrm{mean}(\log(\vn^{(C)}))/2 +
   \mathrm{mean}(\log(\vn^{(K)}))/2
   ).
\end{align*} 
This simplified update scheme is summarized in the rightmost box of  Fig.~\ref{fig:kron_updates}.

\begin{claim}
\label{claim:approx1}
In our Kronecker-based update 
scheme (see Fig.~\eqref{fig:kronecker}),
we have the following identity for each factor $l \in \{C,K\}$. 
\begin{align*}
 \frac{1}{\alpha k^{(l)}}   \mathrm{mean}\big( (\vd^{(l)})^{-1} \odot  \mathrm{diag}(\vW^{(l)})\big) = \frac{1}{m n\alpha} \mathrm{Tr}\big[ (\vS^{(C)})^{-1}\vG  (\vS^{(K)})^{-1}\vG^T \big],
\end{align*} where  $\vS^{(l)}:=\vB^{(l)}\mathrm{Diag}(\vd^{(l)})\big(\vB^{(l)}\big)^T$, $k^{(C)}:=m$, $k^{(K)}:=n$, $\vS^{(C)} \in \mathcal{R}^{ n \times n }$,
and $\vS^{(K)} \in \mathcal{R}^{ m \times m }$.
\end{claim}

\begin{proof}
We will prove the case when $l=C$. Likewise, we can prove the case when $l=K$.
Recall that $\vW^{(C)}$ is defined as $\vW^{(C)}:=(\vB^{(C)})^T \vQ^{(C)} \vB^{(C)}= (\vB^{(C)})^T\vG (\vS^{(K)})^{-1}\vG^T (\vB^{(C)})$. We have the following expression.
\begin{align*}
    \frac{1}{\alpha k^{(l)}}\mathrm{mean}\big[ ( \vd^{(C)}  )^{-1} \odot \mathrm{diag}(\vW^{(C)}) \big] & =  \frac{1}{mn\alpha}\mathrm{Tr}\big[\mathrm{Diag}(\vd^{(C)})^{-1}  \vW^{(C)}\big] \\
    &= \frac{1}{mn\alpha}\mathrm{Tr}\big[
    \mathrm{Diag}(\vd^{(C)})^{-1}
    (\vB^{(C)})^T\vG (\vS^{(K)})^{-1}\vG^T \vB^{(C)}
    \big] \\
    &= \frac{1}{mn\alpha}\mathrm{Tr}\big[
 \underbrace{    \vB^{(C)}  \mathrm{Diag}(\vd^{(C)})^{-1} (\vB^{(C)})^T}_{= ( \vS^{(C)})^{-1} } \vG (\vS^{(K)})^{-1}\vG^T
    \big] 
\end{align*}

\end{proof}

\begin{claim}
\label{claim:approx2}
In our Kronecker-based update 
scheme (see Fig.~\eqref{fig:kronecker}), the updates  of $\alpha$ and $\vd^{(l)}$ can be merged and approximated as 
\begin{align*}
 \alpha \vd^{(l)} \leftarrow
(\alpha \vd^{(l)})\odot \exp(\stepsize_2 \vm^{(l)} ) & \approx \alpha \overbrace{\big[ (1-\gamma \stepsize_2)\vd^{(l)} +  \frac{\stepsize_2}{\alpha k^{(l)}} \mathrm{diag}\big(\vW^{(l)} \big) \big]}^{:= \vn^{(l)}} \\
& =  \alpha \exp\big[\mathrm{mean}( \log(\vn^{(l)}) )\big] \exp\big[\log(\vn^{(l)}) - \mathrm{mean}( \log(\vn^{(l)}) ) \big]
\end{align*} where $l \in \{C,K\}$.
\end{claim}
\begin{proof}
   According to the update scheme in Fig.~\ref{fig:kronecker} and Claim \ref{claim:approx1}, we can reexpress the update in $\alpha$ as
 \begin{align*}
\alpha \leftarrow \alpha \exp( \frac{\stepsize_2}{2 } [\mathrm{mean}(\vm^{(C)}) + \mathrm{mean}(\vm^{(K)})] ) & = \alpha \exp( \frac{\stepsize_2}{2 } [\mathrm{mean}(\vm^{(C)}) + \mathrm{mean}(\vm^{(K)})] ) \\
&= \alpha\exp\big[\frac{\stepsize_2}{2 } \big(-2\gamma + \frac{2}{\alpha mn} \mathrm{Tr}( (\vS^{(C)})^{-1}\vG  (\vS^{(K)})^{-1}\vG^T ) \big)\big] \\
&= \alpha\exp\big[\stepsize_2 \big(-\gamma + \frac{1}{\alpha mn} \mathrm{Tr}( (\vS^{(C)})^{-1}\vG  (\vS^{(K)})^{-1}\vG^T ) \big)\big]
\end{align*} where 
$ \vm^{(l)} = (\vd^{(l)})^{-1} \odot \big[ - \gamma\vd^{(l)} + \frac{1}{ {\alpha} k^{(l)}}   \mathrm{diag}(\vW^{(l)} ) \big] 
$.
Similarly, we can reexpress the update of $\vd^{(l)}$ as
\begin{align*}
\vd^{(l)} & \leftarrow 
\vd^{(l)} \odot \exp\big[ \stepsize_2 [\vm^{(l)} - \mathrm{mean}(\vm^{(l)})] \big] \\
& = \vd^{(l)} \odot  \exp\big[\stepsize_2 \big(  (\vd^{(l)})^{-1} \odot \big( \frac{1}{ {\alpha} k^{(l)}}   \mathrm{diag}(\vW^{(l)} ) \big) - \frac{1}{mn\alpha}\mathrm{Tr}((\vS^{(C)})^{-1}\vG  (\vS^{(K)})^{-1}\vG^T) \big) \big]
\end{align*}
We can merge these updates and truncate the experiential map as
\begin{align*}
    \alpha \vd^{(l)} & \leftarrow \alpha \vd^{(l)} \exp\big[ 
    \stepsize_2 \big( -\gamma + (\vd^{(l)})^{-1} \odot \big( \frac{1}{ {\alpha} k^{(l)}}   \mathrm{diag}(\vW^{(l)} ) \big)  \big)
    \big]  \\
    & \approx 
    \alpha \vd^{(l)} \odot \big( (1-\stepsize_2 \gamma) + (\vd^{(l)})^{-1} \odot \big[ \frac{1}{ {\alpha} k^{(l)}}   \mathrm{diag}(\vW^{(l)} ) \big] \big) \\
    & = \alpha \big[(1-\stepsize_2 \gamma) \vd^{(l)} + \frac{\stepsize_2}{ {\alpha} k^{(l)}}   \mathrm{diag}(\vW^{(l)} )\big]
\end{align*}
\end{proof}

\section{Additional Results for the Empirical Validation}
\label{app:extra_empirical_valida}

\subsection{Full-matrix Update Scheme}
\begin{description}
 \item[Fixed-point matching]\,
 The ground truth in this setting is a fixed-point solution, $\vS_{*} = E[\vg\vg^T]=\vSigma$, to the default update scheme as $\vS_{*}=(1-\stepsize)\vS_{*} + \stepsize \vg_k\vg_k^T $, where
  $\vg_k$ is independently generated from a Gaussian distribution $\vg_k \sim \gauss(\mathbf{0}, \vSigma)$ at each iteration $k$.
We evaluate each scheme at iteration $k$ by comparing its current estimate denoted by $\vS_k^{\text(est)}$ to the fixed point.
We use a relative Frobenius norm 
\scalebox{0.8}{
$\frac{ \|\vS_* - \vS_k^{\text{(est)}}\|_\text{Frob}}{\|\vS_*\|_\text{Frob}}$ }.
and the Wasserstein-2 distance for positive-definite matrices to measure the difference.

\item[\bf Iterate matching]\,
The ground truth is a sequence of matrices $\{\vS_1^{\text{(true)}},\dots,\vS_T^{\text{(true)}}\}$
generated by the default scheme when applying the scheme to the gradient sequence.
We want to match the iterate that the default scheme generates at every step. 
We use a relative Frobenius norm 
\scalebox{0.8}{
$\frac{ \|\vS_k^{\text{(true)}} - \vS_k^{\text{(est)}}\|_\text{Frob}}{\|\vS_k^{\text{(true)}}\|_\text{Frob}}$}.
and the Wasserstein-2 distance to measure the discrepancy between an update scheme and the default update scheme at every iteration $k$.
\end{description}

\subsection{Kronecker-based Update Scheme}
\begin{description}
 \item[Fixed-point matching]\,
 The ground truth is an unstructured fixed-point solution, $\vS_{*} = E[\vg\vg^T]=\vSigma$.
We evaluate a Kronecker-structured scheme in every iteration $k$ by comparing its current
structured estimate to the fixed point. We measure the difference using the same metrics
considered previously.

\item[\bf Iterate matching]
The ground truth is a sequence of unstructured matrices generated by the default
scheme. Our goal is to match the iterate that the default scheme generates using Kronecker
structured approximations We use the same metrics to measure the difference.

\end{description}

\begin{center}
\begin{figure*}
  \centering
 \includegraphics[width=\linewidth]{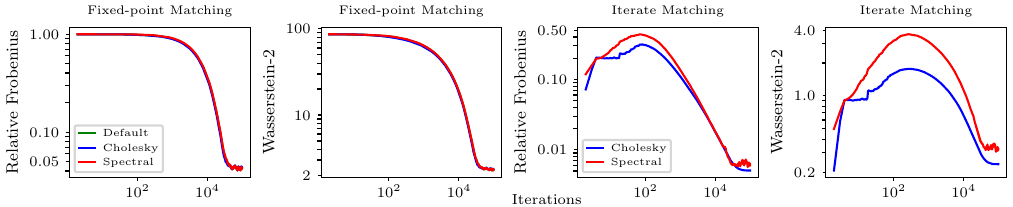}
  \vspace{-2ex}
  \caption{
  Empirical validation of our full-matrix update scheme on estimating preconditioning matrix $\vS \in \real^{100 \times 100}$.
  The first two figures on the left show that our update scheme converges to a fixed-point solution as fast as the default update scheme in $\vS$ and the Cholesky-based scheme. 
  The last two figures illustrate how closely our update scheme matches the iterates generated by the default update scheme at each iteration.
  Our update scheme and the Cholesky-based scheme perform similarly for matching the preconditioner estimates generated by the default scheme.
  }
 \label{fig:full_mat_toy}
\end{figure*}
\end{center}

\begin{center}
\begin{figure*}
  \centering
  \includegraphics[width=\linewidth]{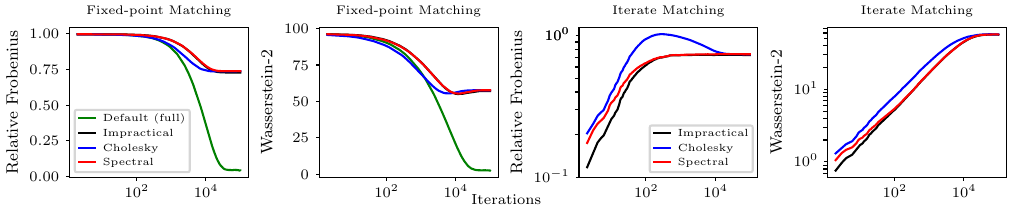}
  \vspace{-2ex}
  \caption{
   Empirical validation of our structured update scheme on estimating a preconditioner $\vS$ using a Kronecker product $\vS^{(C)} \otimes \vS^{(K)}$, where $\vS^{(C)} \in \real^{9 \times 9} $ and $ \vS^{(K)} \in \real^{11 \times 11}$.
  The first two figures on the left show that our update scheme gives a structural approximation of
  a fixed-point solution obtained by the default full-matrix update scheme. %
  Our scheme converges as fast as Kronecker-structured baseline methods, including the impractical projection-based method.
  The last two figures illustrate how closely our  scheme matches the unstructured iterates generated by the default  scheme at each iteration. All update schemes perform similarly due to the structural  approximation  gap.
  }
  \label{fig:kron_toy}
\end{figure*}

\end{center}

\section{Results on Gradient-free Evolutionary Problems}
\label{app:bb_opt}

\begin{figure*}
  \centering
  \includegraphics[width=\linewidth]{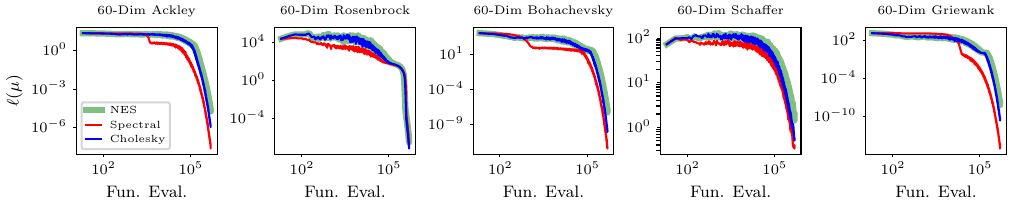}
  \vspace{-2ex}
  \caption{
  Experiments demonstrate the
  efficacy of our update schemes for gradient-free optimization problems defined at Appx.~\ref{app:bb_opt}.
 These problems represent diverse optimization settings, including multimodality and narrow curvature.
  We can see that our method performs similarly to Riemannian baseline methods.
  }
  \label{fig:extra_results}
\end{figure*}

We consider five test problems:
the Ackley function $\ell_\text{Ackley}(\vw)=20-20 \exp(-0.2\sqrt{\frac{1}{d}\sum_{i=1}^d w_i^2}) + e - \exp(\frac{1}{d}\sum_{i=1}^d \cos(2\pi w_i)) $, 
the Rosenbrock function $\ell_\text{Rosenbrock}(\vw)=\sum_{i=1}^{d-1} ( 100 (w_{i+1}-w_i^2)^2 +(w_i-1)^2 )$,
the Bohachevsky function $\ell_\text{Bohachevsky}(\vw)=\sum_{i=1}^{d-1}(w_i^2 + 2 w_{i+1}^2 - 0.3 \cos(3\pi w_i) - 0.4 \cos(4\pi w_{i+1}) + 0.7 )$, the Schaffer function $\ell_\text{Schaffer}(\vw)= \sum_{i=1}^{d-1} (w_i^2 + w_{i+1}^2)^{0.25} [ \sin^2 (50 (w_i^2 + w_{i+1}^2)^{0.1} ) +1 ] $ and the
Griewank function $\ell_\text{Griewank}(\vw)  =\frac{1}{4000} \sum_{i=1}^{d} w_i^2 - \prod_{i=1}^d \cos(\frac{w_i}{\sqrt{i}})+1 $. These problems represent diverse optimization settings, including multimodality and narrow curvature. 
From Fig.~\ref{fig:extra_results}, we can see that our method performs similarly to the baseline methods and demonstrates the efficacy of our update scheme.

\section{Additional Details and Results for Low-precision NN training}

\begin{figure*}
  \centering
 \includegraphics[width=\linewidth]{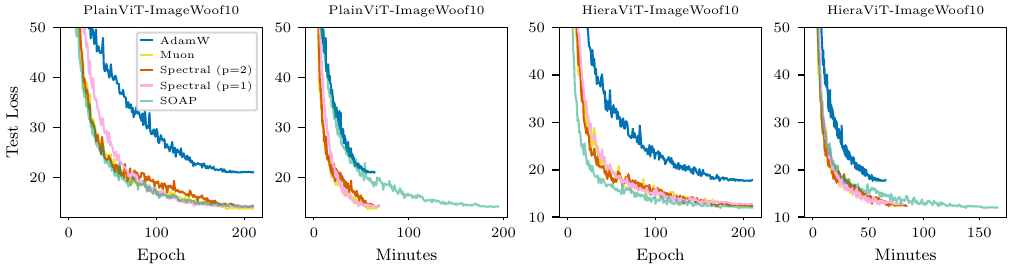}
  \vspace{-2ex}
  \caption{
  Comparison of the methods on the ImageWoof-10 dataset in terms of iteration efficiency and wall-clock time.
  These plots demonstrate the efficiency of our update schemes for low-precision NN training compared to state-of-the-art methods such as Muon and SOAP. 
  }
 \label{fig:imagewoof}
\end{figure*}

\begin{figure*}
  \centering
 \includegraphics[width=\linewidth]{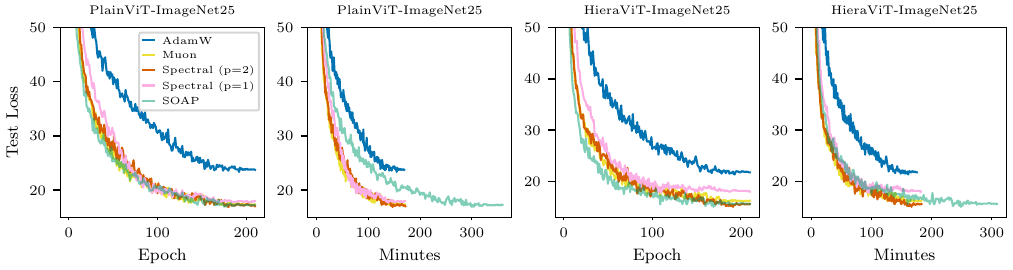}
  \vspace{-2ex}
  \caption{
  Comparison of the methods on the ImageNet-25 dataset in terms of iteration efficiency and wall-clock time.
  These plots demonstrate the efficiency of our update schemes for low-precision NN training compared to state-of-the-art methods such as Muon and SOAP. 
  }
 \label{fig:imagenet25}
\end{figure*}

We consider two vision transformer-based (ViT) models to demonstrate the effectiveness of our methods in half precision.
We train ViTs for 210 epochs with mini-batch size 128 for ImageWoof-10 and 256 for ImageNet-25. 
The ImageWoof-10 dataset is obtained from \url{https://github.com/fastai/imagenette}. The ImageNet-25 dataset is a subset of the ImageNet-1k dataset by randomly sampling 25 classes from the 1000 classes.
We use data augmentation techniques such as MixUp 
\citep{zhang2017mixup} and CutMix \citep{yun2019cutmix} during training. 
We use a cosine learning rate schedule suggested by \citet{chen2023symbolic}.
We use PyTorch’s built-in AdamW and official implementation for SOAP (\url{https://github.com/nikhilvyas/SOAP}) and Muon (\url{https://github.com/KellerJordan/modded-nanogpt}).
For SOAP and our method, we update their matrix preconditioners at each 2 iterations.
We tune each optimizer's available hyperparameters (HPs) using
random search \citep{choi2019empirical}. We use 80 runs for the search.
We use Pytorch's mixed precision with BFP-16 for NN training.
All methods except SOAP can use
BFP-16 data types and perform updates in BFP-16. SOAP has to internally cast BFP-16 gradients into FP-32 and performs matrix decomposition  (e.g., QR and eigendecomposition) in FP-32.
This is because matrix decomposition is numerically unstable in half-precision and, therefore, unsupported/unavailable in PyTorch, JAX, Julia, Scipy/Numpy, and LAPACK.
Figs.~\ref{fig:imagewoof} and \ref{fig:imagenet25} illustrate the performance of our Kronecker-based update scheme.

\label{app:extra_nn_training}
\begin{figure*}
  \centering
 \includegraphics[width=\linewidth]{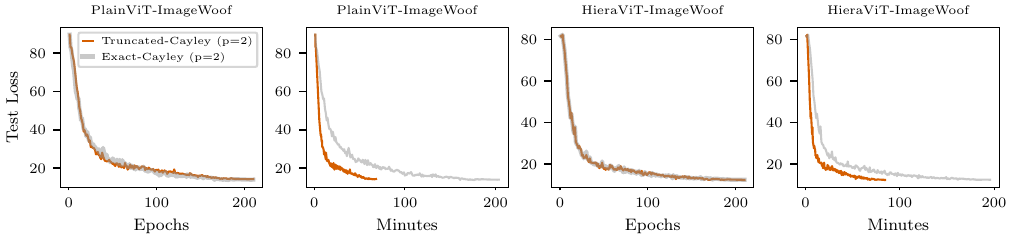}
  \vspace{-2ex}
  \caption{
  Comparison of our Kronecker-based update schemes with $p=2$ (see the bottom box of Fig.~\ref{fig:kronecker}) using the exact Cayley and the truncated Cayley maps.
  We use nonconstant $\stepsize_2^{(l)}:=\frac{ \bar{ \stepsize}_2 \alpha \kappa^{(l)} }{ \| 
  \mathrm{Skew}( \mathrm{Tril}(\vU^{(l)})  ) 
 \|_{\text{Frob}} }$  and use the same hyperparameters in both cases, where $\kappa^{(C)}=m$, $\kappa^{(K)}=n$, $\vU^{(C)} \in \mathcal{R}^{n \times n}$, and  $\vU^{(K)} \in \mathcal{R}^{m \times m}$. Note that $\bar{\stepsize_2}$ is a constant hyperparameter.
We can see that both maps work equally well in terms of iterations/epochs. However, using the exact Cayley map is much slower regarding running time because it uses matrix inversion.
  }
 \label{fig:exact_vs_truncated}
\end{figure*}

\section{
Handling Redundancy due to Repeated Entries of $\vd$ }
\label{app:handling_repated_d}

Recall that eigendecomposition is not unique when having repeated eigenvalues.
Our spectral parametrization is also not unique when $\vd$ has duplicate entries. 
In this case, the Fisher-Rao metric (coordinate representation) is singular.
We allow $\vd$ to have repeated entries and address the singularity using the Moore–Penrose inversion \citep{van2023invariance}. %
Computing this inversion is easy because we diagonalize the metric at evaluation points. 
We also use this inversion to improve numerical stability if $\vd$ has very close entries.

\section{Proof of Claim \ref{claim:linear_invariance}}
\label{app:proof_lemma_linear}
{\bf (Formal) Claim 1}
Consider an unconstrained parameterization $\vtau \in \mathcal{R}^d$ and a linear reparametrization $\vlambda \in \mathcal{R}^p$ so that  $\vtau:= \vU \vlambda $, where 
the metric $\vF_\tau$ in coordinate $\vtau$ is nonsingular (and thus SPD) everywhere, $\vU \in \mathcal{R}^{d \times p}$ is a constant matrix with rank $d$, and $p \geq d$. Then performing (unconstrained) RGD (see \eqref{eq:rgd}) in coordinate $\vtau$ is equivalent 
to performing RGD in coordinate $\vlambda$ (i.e., $\vtau_k = \vU \vlambda_k$ for each iteration $k$)

{\bf Remark 1:}
If a linear reparametrization is written as $\vlambda = \vA \vtau$, we can define $\vU$ as follows, where $\vA \in \mathcal{R}^{p \times d}$ with rank $d$.
We define $\vU := \vV \begin{bmatrix}
    \vD^{-1} & \mathbf{0}
\end{bmatrix} \vK^T$, where  $\vV$, $\vD$, and $\vK$ are obtained via
the singular value decomposition of $\vA$ so that $\vA = \vK \begin{bmatrix}
    \vD \\ \mathbf{0}
\end{bmatrix}  \vV^T$. 
Recall that $\vK \in \mathcal{R}^{p \times p}$ and $\vV \in \mathcal{R}^{d \times d} $ are orthogonal matrices. Moreover, $\vD \in \mathcal{R}^{d \times d}$ is a non-singular diagonal matrix because of the rank of $\vA$.
It is easy to verify that
$\vU$ is well-defined because 
\begin{align*}
\vU \vlambda & = \vU \vA \vtau \\
   & =\vV \begin{bmatrix}
    \vD^{-1} & \mathbf{0}
\end{bmatrix} \underbrace{\vK^T 
 \vK}_{=\vI_p} \begin{bmatrix}
    \vD \\ \mathbf{0}
\end{bmatrix}  \vV^T \vtau \\
&= \underbrace{\vV \vI_d \vV^T}_{=\vI_d} \vtau = \vtau.
\end{align*}

{\bf Remark 2:}
Recall that by the definition of the Riemannian metric, we have $\vF_\lambda = \vU^T \vF_\tau \vU$.
Moreover, we have $\vg_\tau:=\nabla_\tau \mathcal{L}$ and $\vg_\lambda = \nabla_\lambda
\mathcal{L}= \vU^T \vg_\tau$ by the chain rule.
Note that when $p > d$, $\vlambda$ is in a linear over-parametrized space.
The claim implies that the over-parametrized $\vlambda$ preserves the invariance when $\vtau = \vU\vlambda$.
Notably, we can see that the metric $\vF_\lambda \in \mathcal{R}^{p \times p}$ in coordinate $\vlambda$ is singular when $p > d$. This is due to over-parametrization.

\begin{proof}
We will use proof by induction. 
We initialize $\vlambda$ so that $\vtau_0 = \vU\vlambda_0$. Thus, the base case holds.
At iteration $k$, by induction, we have $\vtau_k = \vU \vlambda_k  $. Now, we show that this relationship holds at iteration $k+1$.

By \eqref{eq:rgd}, $\tau_{k+1}$ is updated by following this expression:
\begin{align*}
    \vtau_{k+1} = \vtau_k -\stepsize [ \vF_\tau (\vtau_k)]^{-1} \vg_\tau(\vtau_k) 
\end{align*}

Since $\vF_\lambda$ is singular, we have to use  the Moore-Penrose inverse and it is computed as
\begin{align*}
    [\vF_\lambda]^{-1} &=[ \vU^T  \vF_\tau \vU]^{-1} \\
    &=[ \underbrace{\vU^T  \vL}_{:=\vC} 
 \underbrace{\vL^T \vU}_{:=\vD}]^{-1} \\
 &= \vD^T (\vD\vD^T)^{-1}(\vC\vC^T)^{-1}\vC^T \,\,\,\text{(according to Eq. 224 of \citet{petersen2008matrix})}\\
 &= \vU^T \vL (\vL^T\vU\vU^T\vL)^{-1}(\vL^T\vU\vU^T\vL)^{-1} \vL^T \vU \\
 &= \vU^T (\vU\vU^T)^{-1} \vF_\tau^{-1} (\vU\vU^T)^{-1} \vU,
\end{align*} where we use the Cholesky decomposition for $\vF_\tau = \vL\vL^T$ since $\vF_\tau$ is SPD.

Likewise, $\lambda_{k+1}$ is updated by RGD.
We have the following this expression:
\begin{align*}
\vU \{ 
    \vlambda_{k+1} \} &= \vU\{\underbrace{ \vlambda_k -\beta [ \vF_\lambda (\vlambda_k)]^{-1} \vg_\lambda(\vlambda_k)}_{\text{by the definition of RGD}} \} \\
& =\vU\{ \vlambda_k -\stepsize[ \vF_\lambda (\vlambda_k)]^{-1} \big(\underbrace{\vU^T \vg_\tau(\vtau_k)}_{\text{by the chain rule}}\big) \} \\
   & = \vU \{ \vlambda_k -\stepsize \big[ \vU^T (\vU\vU^T)^{-1} \vF_\tau^{-1} (\vU\vU^T)^{-1} \vU  \big] \big( \vU^T \vg_\tau(\vtau_k)\big)\}
 \,\,\,\text{(using the Moore-Penrose inverse)} \\
 &= \vU\{ \vlambda_k - \stepsize \vU^T (\vU\vU^T)^{-1} \vF_\tau^{-1} \vg_\tau(\vtau_k)\} \\
 &= \underbrace{ \vU \vlambda_k}_{=\vtau_k} - \stepsize 
 \vF_\tau^{-1} \vg_\tau(\vtau_k) \,\,\, \text{(by assumption)} \\
 &= \vtau_{k+1}
\end{align*}

Thus, by induction, we show the claim holds.
\end{proof}

\section{Proof of Claim \ref{claim:eigen_full_mat_case}}
\label{app:proof_lemma_full}
{\bf (Formal) Claim 2}
Our update scheme for $\vS$ in the top box of Fig~\ref{fig:kronecker} is equivalent to the scheme in Eq.~\eqref{eq:root_free} up to first-order in terms of $\stepsize_2$ (i.e., $\vB_k \mathrm{Diag}(\vd_k)\vB_k^T = \vS_k + O(\stepsize_2^2)$) when $\vd$ does not have repeated entries and using the same sequence of GOPs $\{\mathcal{H}_1, \dots, \mathcal{H}_T \}$, where $\mathcal{H}_k:=\vg_k\vg_k^T$ is a GOP.

\begin{proof}

We will use proof by induction.
When $k=0$, we initialize  $\vB$ and $\vd$, and define $\bar{\vS}_0$ so that $\bar{\vS}_0 := \vB_0\mathrm{Diag}(\vd_0)\vB_0^T = \vS_0$. Thus, the base case holds.

By induction, the claim holds at iteration $k$. That is $ \bar{\vS}_k:=\vB_k \mathrm{Diag}(\vd_k)\vB_k^T =\vS_k+O(\stepsize_2)$
We want to show that the claim also holds $k+1$ at iteration.  
Recall that we update the spectral factors using the following rule at iteration $k$.
\begin{align}
     \vd_{k+1} & \leftarrow  \vd_k \odot \exp\{ \stepsize_2\, \vd_k^{-1} \odot [ -\gamma \vd_k +    \mathrm{diag}(\vB_k^T  \mathcal{H}_k \vB_k )  ]  \} \nonumber \\
  \vB_{k+1} & \leftarrow \vB_k\mathrm{Cayley}(\frac{\stepsize_2}{2}\, \mathrm{Skew}({\mathrm{Tril}(\vU)}) ), 
\end{align} where $\mathcal{H}_k=\vg_k\vg_k^T$ and 
the $(i,j)$-th entry of $\vU$ is $[U]_{ij}:= -[\vB_k^T \mathcal{H}_k \vB_k]_{ij}/(d_i - d_j) $, where $\vd$ has no repeated entries by our assumption.
We want to show that the above update scheme is equivalent to the default update scheme in \eqref{eq:root_free} up to first order in terms of $O(\beta_2)$. 
\begin{align}
    \vS_{k+1} \leftarrow  (1-\stepsize_2\gamma) \vS_k + \stepsize_2 \underbrace{\mathcal{H}_k}_{ =\vg_k \vg_k^T}
\end{align}

Let $\vQ_k := \vB_k^T  \mathcal{H}_k \vB_k$.
Recall that the Cayley map is defined as $\mathrm{Cayley}(\vN)=(\vI +\vN)(\vI-\vN)^{-1}$.
Using the first-order approximation of $(\vI-\vN)^{-1}$, we have $(\vI-\stepsize_2\vN)^{-1} = \vI +\stepsize_2 \vN + O(\stepsize_2^2) $.
Thus, we have $\vB_{k+1} =\vB_k \mathrm{Cayley}(\frac{\stepsize_2}{2}\, \mathrm{Skew}({\mathrm{Tril}(\vU)}) = \vB_k (\vI+\frac{\stepsize_2}{2}\vN) (\vI+\frac{\stepsize_2}{2}\vN + O(\stepsize_2^2))= \vB_k (\vI+ \stepsize_2 \vN + O(\stepsize_2^2) )$, where $\vN:=\mathrm{Skew}({\mathrm{Tril}(\vU)})$.
Similarly, we have
$\vd_{k+1} = \vd_k \odot [ 1 + \stepsize_2 \vd_k^{-1} \odot (-\gamma \vd_k + \mathrm{diag}(\vQ_k) ) + O(\stepsize_2^2)] = \vd_k + \stepsize_2 \vw_k+O(\stepsize_2^2)$
by using the first-order truncation of the exponential map, where 
$\vw_k:=-\gamma \vd_k+\mathrm{diag}(\vQ_k)$.
Notice that
\begin{align*}
    &\bar{\vS}_{k+1} \\ 
    :=& \mathbf{B}_{k+1} \mathrm{Diag}(\vd_{k+1}) \mathbf{B}_{k+1}^{T} \\
    = &\mathbf{B}_{k} \Big[ (\vI+\stepsize_2 \vN + O(\stepsize_2^2)) \mathrm{Diag}( \vd_k +\stepsize_2 \vw_k + O(\stepsize_2^2)) \Big]
    (\vI+\stepsize_2 \vN + O(\stepsize_2^2))^T 
 \vB_k^T \\
    =& \vB_k \Big[ \vD_k  +\stepsize_2 \vN \vD_k + \stepsize_2\vW_k + O(\stepsize_2^2) \Big] (\vI+\stepsize_2 \vN + O(\stepsize_2^2))^T\vB_k^T\\
    =& \vB_k \Big[ \vD_k  +\stepsize_2 \vN \vD_k + \stepsize_2\vW_k + \stepsize_2 \vD_k \vN^T + O(\stepsize_2^2) \Big]\vB_k^T \\
=& \vB_k \vD_k \vB_k^T + \stepsize_2 \vB_k ( \vN\vD_k + \vW_k + \vD_k\vN^T) \vB_k^T + O(\stepsize_2^2) \\
=& \bar{\vS}_k + \stepsize_2 \vB_k (\vN\vD_k + \vD_k\vN^T) \vB_k^T + \stepsize_2 \vB_k \vW_k \vB_k^T +O(\stepsize_2^2), \, (\vN \text{ is skew-symmetric})\\
=&\bar{\vS}_k + \stepsize_2 \vB_k (\vN\vD_k - \vD_k\vN) \vB_k^T + \stepsize_2 \vB_k \vW_k \vB_k^T+O(\stepsize_2^2)
\end{align*} where $\vD_k:=\mathrm{Diag}(\vd_k)$ and $\vW_k:=\mathrm{Diag}(\vw_k)$

Observation (1): Since $\vW_k=\mathrm{Diag}(-\gamma \vd_k + \mathrm{diag}(\vB_k^T \mathcal{H}_k\vB_k))=-\gamma \vD_k+  \mathrm{Diag}( \mathrm{diag}(\vB_k^T \mathcal{H}_k \vB_k) ) $, we have
\begin{align}
   \vB_k \vW_k \vB_k^T = -\gamma \vB_k \vD_k \vB_k^T + \vB_k \mathrm{Ddiag}(\vQ_k) \vB_k^T
\end{align} where $ \mathrm{Ddiag}(\vQ_k) $ denotes the  diagonal part of $\vQ_k=\vB_k^T \mathcal{H}_k \vB_k$

Observation (2): Since $\vN = \mathrm{Skew}(\mathrm{Tril}(\vU))$ and $[\vU]_{ij}=  -[\vB_k^T \mathcal{H} \vB_k]_{ij}/(d_i - d_j)=-[\vQ_k]_{ij}/(d_i-d_j)$, we can show that $\vN\vD_k - \vD_k\vN$ is indeed a symmetric matrix with zero-diagonal entries.
Moreover,  
the low-triangular half ($i>j$) of the matrix can be expressed as
\begin{align}
    [\vN \vD_k -\vD_k \vN]_{ij} = (d_j-d_i) [\vU]_{ij} = [\vQ_k]_{ij}.
\end{align} where $d_j \neq d_i$ since $\vd$ has no repeated entries.
Thus, we have 
$\vN\vD_k - \vD_k\vN = \vQ_k - \mathrm{Ddiag}(\vQ_k)$.

Using Observations (1) and (2), we have
\begin{align*}
    &\bar{\vS}_{k+1} \\ 
    =& \vB_k \Big[ \vD_k  +\stepsize_2 \vN \vD_k + \stepsize_2\vW_k + \stepsize_2 \vD_k \vN^T + O(\stepsize_2^2) \Big]\vB_k^T \\
    =& \vB_k\vD_k\vB_k^T +\stepsize_2 \Big[ \vB_k \Big(\vQ_k-\mathrm{Ddiag}(\vQ_k)\Big) \vB_k^T  -\gamma \vB_k\vD_k\vB_k^T + \vB_k \mathrm{Ddiag}(\vQ_k) \vB_k^T \Big] + O(\stepsize_2^2) \\
    =& (1-\stepsize_2\gamma)  \vB_k\vD_k\vB_k^T + \stepsize_2 \vB_k \Big[ \vQ_k \Big] \vB_k^T + O(\stepsize_2^2),\, (\text{Note: } \vQ_k=\vB_k^T \mathcal{H}_k\vB_k)\\
    =& (1-\stepsize_2\gamma)\bar{\vS}_k + \stepsize_2 \mathcal{H}_k + O(\stepsize_2^2) = (1-\stepsize_2\gamma)\vS_k + \stepsize_2 \mathcal{H}_k + O(\stepsize_2^2).
\end{align*}  
By induction, we can show the claim holds.
\end{proof}

\section{Proof of Claim 
\ref{claim:eig_constraint2}}
{\bf Claim 3}
The map in \eqref{eq:translation_map} satisfies the constraints in \eqref{eq:eig_opt}.

\begin{proof}
    
Now, we show the map satisfies the spectral parameter constraints in \eqref{eq:eig_opt}.
Since $\vmu$ is unconstrained, we only consider the update on $\vd$ and $\vB$.

Recall that the current point  $(\vd_k,\vB_k)$ is in the spectral coordinate. Thus, we have $\vd_k >0$ and $\vB_k$ is orthogonal.
According to the map \eqref{eq:translation_map}, it is easy to see that $\vd(\vm)=\vd_k \odot \exp(\vm) >0$ because $\vd_k>0$. Thus, $\vd(\vm)$ satisfies the parameter constraints.
Now, we show that $\vB(\vM)$ is also orthogonal. 
Notice that a product of two orthogonal matrices is also orthogonal. Because $\vB_k$ is orthogonal, we only need to show that
the output of the Cayley map, $\mathrm{Cayley}(\mathrm{Skew}({\color{red} \mathrm{Tril}(\vM)}))$,  is orthogonal. 
Let $\vN:=\mathrm{Skew}({\color{red} \mathrm{Tril}(\vM)})$. We know that $\vN$ is skew-symmetric.
We can verify that the Cayley map satisfies the orthogonal constraint.
Consider the following expression:
\begin{align*}
\big( \mathrm{Cayley}(\vN) \big)^T \mathrm{Cayley}(\vN) & =   (\vI - \vN)^{-T} (\vI+\vN)^T (\vI+\vN) (\vI - \vN)^{-1} \\
&=  (\vI - \vN)^{-T} (\vI-\vN) (\vI+\vN) (\vI - \vN)^{-1}  \\
&= (\vI - \vN)^{-T} (\vI+\vN) (\vI-\vN) (\vI - \vN)^{-1} \\
&= (\vI - \vN)^{-T} (\vI-\vN)^{T} (\vI-\vN) (\vI - \vN)^{-1}  = \vI
\end{align*} where we use the fact that $\vN$ is skew-symmetric such as $\vN^T=-\vN$.

Likewise, we can show 
$\mathrm{Cayley}(\vN)  \big( \mathrm{Cayley}(\vN) \big)^T =\vI$. 
Thus, the output of the Cayley map is a square orthogonal matrix.

\end{proof}

\section{Proof of Claim 
\ref{claim:fisher_full}
}
\label{app:claim_fisher_full}

{\bf Claim 5}
    The exact Fisher-Rao metric $\vF_\eta(\veta_\text{cur})$ (for a full-matrix Gaussian) evaluated at the origin \scalebox{0.8}{ $\veta_\text{cur}\equiv\mathbf{0}$} is \emph{diagonal} and has the following  closed-form expression:
\begin{align}
    \vF_\eta(\vm, \mathrm{vecTril}(\vM), \vdelta)\big|_{ \vdelta=\mathbf{0}, \vm=\mathbf{0}, \vM =\mathbf{0} } = \begin{bmatrix}
          \vF_{mm} & \mathbf{0} &\mathbf{0}  \\
         \mathbf{0}  & \vF_{MM}& \mathbf{0} \\
     \mathbf{0} & \mathbf{0} & \vF_{\delta\delta}   \\
    \end{bmatrix},
    \label{eq:fim_full_local}
\end{align} 
where  $\vF_{\delta\delta}=\vI$,
$\vF_{mm}=\half \vI$,
$\vF_{MM}= \mathrm{Diag}( \mathrm{vecTril}(\vC) )$, $\mathrm{vecTril}(\vC)$ represents the vectorization of the low-triangular half of $\vC$ excluding diagonal entries and its $(i,j)$-th entry is $[\vC]_{ij}=4( \frac{d_i}{d_j} + \frac{d_j}{d_i} -2 )\geq 0$ and $d_i$ denotes the $i$-th entry of $\vd_k$.
The metric becomes singular when $\vd$ has repeated entries (i.e., $d_i =d_j$ for $i \neq j$). In this case, we use the Moore-Penrose inversion to inverse the metric.

To verify this statement, we can analytically compute the Fisher-Rao metric according to the following expression.  
\begin{align*}
    \vF_\eta(\veta) &= E_{w\sim q}[\nabla_\eta \log q(\vw;\veta)\nabla_\eta^\top \log q(\vw;\veta)]  \\
    &= - E_{w\sim q}[\nabla_\eta^2 \log q(\vw;\veta)],
\end{align*} where we use the expression in the last line to compute the metric and obtain the simplified expression in Eq.~\eqref{eq:fim_full_local}.

\section{Proof of Claim 
\ref{claim:kron_unique}
}
\label{app:claim6_proof}
{\bf Claim 6}
    A Kronecker-structured positive-definite matrix $\vS$ can be uniquely expressed as  
    \scalebox{0.8}{ $\vS=\alpha [\vS^{(C)} \otimes \vS^{(K)}]$ } with constraints  \scalebox{0.8}{$\mathrm{det}(\vS^{(C)})=\mathrm{det}(\vS^{(K)})=1$ } and \scalebox{0.8}{$\alpha>0$}.

\begin{proof}
We will show that 
$\vS_1:=\alpha_1 [\vS_1^{(C)} \otimes \vS_1^{(K)}] =\vS_2:= \alpha_2 [\vS_2^{(C)} \otimes \vS_2^{(K)}] $ implies $\alpha_1=\alpha_2$,
and $\vS_1^{(l)}=\vS_2^{(l)}$ for $l \in \{C,K\}$, 
where $\vS_1^{(l)}$ and $\vS_2^{(l)}$ are assumed to have the same shape.

Thanks to the determinant constraint, we have
$\alpha_1=\alpha_2$ because of $\mathrm{det}(\vS_1)=\mathrm{det}(\vS_2)$. 

Since $\vS_1=\vS_2$ is SPD, we have
\begin{align*}
    \vI & = \vS_1^{-1} \vS_2 \\
   &=  \{ (\vS_1^{(C)})^{-1} \vS_2^{(C)} \} \otimes
    \{ (\vS_1^{(K)})^{-1} \vS_2^{(K)} \}
\end{align*}

Notice that 
$ \mathrm{det} ( (\vS_1^{(l)})^{-1} \vS_2^{(l)} ) = 1
$ due to the determinant constraint for $l \in \{C,K\}$.
By the definition of the Kronecker structure, the above expression only holds when  $(\vS_1^{(K)})^{-1} \vS_2^{(K)}=\vI^{(K)}$ and  $(\vS_1^{(C)})^{-1} \vS_2^{(C)}=\vI^{(C)}$ for regardless of the shapes of $\vS^{(C)}$ and $\vS^{(K)}$.

Thus, we have $\vS_1^{(l)}=\vS_2^{(l)}$ for $l \in \{C,K\}$.
\end{proof}

\begin{comment}
\section{Proof of Claim 
\ref{claim:local_kron_unique}
}
{\bf Claim 7}
The transformation map (see Eq.~\eqref{eq:kron_trans_map}) is one-to-one if $\vd^{(l)}$ has no repeated entries for each Kronecerk factor $l$.

\begin{proof}

It is easy to see that the map is injective for the local coordinate for $\alpha$.
The remaining proof is similar to the proof of Claim~\ref{claim:eig_constraint} in Appx.~\ref{app:claim4_proof}. 
\end{proof}
\end{comment}

\section{Proof of Claim 
\ref{claim:fisher_kron}
}
\label{app:kron_fim}
In a Kronecker case, we consider
this spectral factorization $\vS = \alpha [ (\vB^{(C)}\mathrm{Diag}(\vd^{(C)})(\vB^{(C)})^T) \otimes (\vB^{(K)}\mathrm{Diag}(\vd^{(K)})(\vB^{(K)})^T) ]$.
At iteration $k$, we create a local coordinate $\veta:=(\vdelta,n, \vm^{(C)},\vM^{(C)}, \vm^{(K)},\vM^{(K)})$ at the current point $\vtau_k:=(\vmu_k,\alpha_k, \vd_k^{(C)}, \vB_k^{(C)},\vd_k^{(K)}, \vB_k^{(K)})$ and use this local transformation map
\begin{align}
\vtau(\veta;\vtau_k) :=
\begin{bmatrix}
\alpha(n;\vtau_k) \\
\vd^{(C)} (\vm^{(C)};\vtau_k)  \\
\vB^{(C)} (\vM^{(C)};\vtau_k) \\
\vd^{(K)} (\vm^{(K)};\vtau_k)  \\
\vB^{(K)} (\vM^{(K)};\vtau_k) \\
\vmu (\vdelta;\vtau_k) 
\end{bmatrix} = \begin{bmatrix}
    \alpha_{k}  \exp( n ) \\
    \vd_{k}^{(C)} \odot \exp( \vm^{(C)} ) \\
\vB_{k}^{(C)} \mathrm{Cayley}(\mathrm{Skew}(\mathrm{Tril}(\vM^{(C)}))) \\
    \vd_{k}^{(K)} \odot \exp( \vm^{(K)} ) \\
\vB_{k}^{(K)} \mathrm{Cayley}(\mathrm{Skew}(\mathrm{Tril}(\vM^{(K)}))) \\
    \vmu_k + (\vB_k^{(C)}\otimes \vB_k^{(K)}) ( \mathrm{Diag} (\vd_k^{(C)}) \otimes \mathrm{Diag}( \vd_k^{(K)}) )^{-1/2} \vdelta 
\end{bmatrix},
\label{eq:kron_trans_map}
\end{align} where $\vm^{(C)}=[m_1^{(C)}, m_2^{(C)}, \dots, m_{l-1}^{(C)}, -\sum_{i}^{l-1} m_i^{(C)}]$ has $l$ entries but only $(l-1)$ free variables since $\sum(\vm^{(C)})=0$.

To verify this statement, we can analytically compute the Fisher-Rao metric according to its definition. 

\begin{align}
 &   \vF_\eta(n, \mathrm{Free}(\vm^{(C)}), \mathrm{vecTril}(\vM^{(C)}), \mathrm{Free}(\vm^{(K)}), \mathrm{vecTril}(\vM^{(k)}), \vdelta
 )\big|_{ \veta=\mathbf{0}} \\
 = & \begin{bmatrix}
      \vF_{\alpha\alpha} & \mathbf{0} & \mathbf{0} & \mathbf{0}& \mathbf{0} & \mathbf{0}\\
          \mathbf{0} & \vF_{m^{(C)}m^{(C)}} & \mathbf{0} & \mathbf{0}& \mathbf{0} & \mathbf{0} \\
         \mathbf{0}  & \mathbf{0}& \vF_{M^{(C)}M^{(C)}}& \mathbf{0}& \mathbf{0} & \mathbf{0} \\
         \mathbf{0}  & \mathbf{0} & \mathbf{0} & \vF_{m^{(K)}m^{(K)}}& \mathbf{0} & \mathbf{0}\\
         \mathbf{0}  & \mathbf{0} & \mathbf{0} & \mathbf{0}& \vF_{M^{(K)}M^{(K)}} & \mathbf{0}\\
     \mathbf{0} & \mathbf{0} & \mathbf{0} & \mathbf{0}& \mathbf{0} & \vF_{\delta\delta}
    \end{bmatrix}
    \label{eq:fim_kron_local}
\end{align} where
$\mathrm{vecTril}(\vC)$ represents the low-triangular half of $\vC$ excluding diagonal entries and $\mathrm{Free}(\vm)$ extracts free variables from $\vm$.

We can see that the Fisher-Rao is block diagonal with six blocks.

The first two blocks are
$\vF_{\delta\delta}=\vI$ and $\vF_{\alpha\alpha}=\half$.
For each Kronecker factor, we have two blocks. 
For notation simplicity, 
we drop the factor index $C$ in $\vF_{m^{(C)}m^{(C)}}$ and $\vF_{M^{(C)}M^{(C)}}$.

For each Kronecker factor,  
$\vF_{MM}= \mathrm{Diag}( \mathrm{vecTril}(\vW) )$, $\mathrm{vecTril}(\vW)$ represents the low-triangular half of $\vW$ excluding diagonal entries and its $(i,j)$-th entry is $[W]_{ij}=4( \frac{d_i}{d_j} + \frac{d_j}{d_i} -2 )\geq 0$ and $d_i$ denotes the $i$-th entry of $\vd_k$ for the factor.
The $\vF_{mm}$ is non-diagonal but its inverse can be computed as 
$\vF_{mm}^{-1}= 2
\begin{bmatrix}
\frac{l-1}{l} & \frac{1}{l} & \dots & \frac{1}{l}  \\
\frac{1}{l} & \frac{l-1}{l} & \dots & \frac{1}{l} \\
\cdot & \cdot & \dots & \cdot \\
\frac{1}{l} & \frac{1}{l} & \dots & \frac{l-1}{l} 
\end{bmatrix}  \in \real^{(l-1) \times (l-1)}
$ for the $(l-1)$ free variables in $\vm$ denoted by $\mathrm{Free}(\vm)$.
Furthermore, the natural-gradient w.r.t. $\vm$ can also be simplified. 

\section{Complete Derivation for the Full-matrix Update}
\label{sec:deri_full}
According to Claim \ref{claim:fisher_full},  the Fisher-Rao metric under this  local coordinate system is  diagonal as
\begin{align}
    \vF_\eta(\vm, \mathrm{vecTril}(\vM),\vdelta)\big|_{ \vdelta=\mathbf{0}, \vm=\mathbf{0}, \vM =\mathbf{0} } = \begin{bmatrix}
          \vF_{mm} & \mathbf{0} &\mathbf{0} \\
         \mathbf{0}  & \vF_{MM} & \mathbf{0}\\
      \mathbf{0} & \mathbf{0} & \vF_{\delta\delta}  \\
    \end{bmatrix}
\end{align} where  $\vF_{\delta\delta}=\vI$,
$\vF_{mm}=\half \vI$,
$\vF_{MM}= \mathrm{Diag}( \mathrm{vecTril}(\vC) )$, $\mathrm{vecTril}(\vC)$ represents the low-triangular half of $\vC$ excluding diagonal entries and its $(i,j)$-th entry is $[C]_{ij}=4( \frac{d_i}{d_j} + \frac{d_j}{d_i} -2 )\geq 0$ and $d_i$ denotes the $i$-th entry of $\vd_k$.

Use the approximation  in Eq.~\eqref{eq:delta_approx}
 \begin{align}
    \vg_\mu & :=   \partial_\mu \mathcal{L} \mystein E_{w \sim q}[ \nabla_w \ell] \mydel \nabla_\mu \ell =\vg\\
       2\vg_{S^{-1}} &:= 2\partial_{S^{-1}}\mathcal{L} \mystein E_{w\sim q}[\nabla_w^2 \ell ] - \vS \mydel \nabla_\mu^2 \ell -\vS  \approx\mathcal{H} -\vS.
    \end{align}
 where $\vg := \nabla_\mu \ell(\vmu)$ is the gradient of $\ell$, $\mathcal{H}:= \vg\vg^T$ is a Hessian approximation. 

The Euclidean gradient w.r.t local coordinate $(\vdelta,\vm,\vM)$ are
\begin{align}
 \vg_\delta \big|_{\delta=0} &= \vD_k^{-1/2} \vB_k^T \vg_\mu \\
    \vg_m \big|_{m=0} &= - \vd_k^{-1} \odot \mathrm{diag}(\vB_k^T \vg_{S^{-1}} \vB_k) \\
     \vg_{\mathrm{vecTril}(M)} \big|_{M=0} &= 4 \mathrm{vecTril}(\vB_k^T \vg_{S^{-1}} \vB_k \vD_k^{-1} - \vD_k^{-1} \vB_k^T  \vg_{S^{-1}} \vB_k)
\end{align} where $\vD_k:= \mathrm{Diag}(\vd_k)$. Recall that we use a gradient outer product $\mathcal{H}=\vg\vg^T$ as a Hessian approximation. %

The metric can still be singular 
when $\vd$ has repeated entries (i.e., $d_i = d_j$ for $i \neq j$) since $\vF_{MM}$ can be singular.
We can use the Moore-Penrose inverse when computing the inverse.
Thanks to this coordinate system,
$\vF_{MM}$ is indeed a diagonal matrix.

Thus, we can simplify the RGD update as
\begin{align}
     \begin{bmatrix}
        \vm \\
        \mathrm{vecTril}(\vM) \\
        \vdelta 
    \end{bmatrix} \leftarrow
       \begin{bmatrix}
         \mathbf{0} - \stepsize_2 \vF_{mm}^{-1} \vg_m \big|_{m=0} \\
         \mathbf{0} - \stepsize_2 \vF_{MM}^{-1}\vg_{\mathrm{vecTril}(M)} \big|_{M=0} \\
        \mathbf{0}  - \stepsize_1  \vF_{\mu\mu}^{-1}\vg_\delta 
    \end{bmatrix} =
      \begin{bmatrix}
         \mathbf{0} - 2 \stepsize_2 \vg_m \big|_{m=0} \\
         \mathbf{0} - \stepsize_2 \vF_{MM}^{-1}\vg_{\mathrm{vecTril}(M)} \big|_{M=0} \\ 
        \mathbf{0} - \stepsize_1 \mathrm{Diag}(\vd)^{-1/2} \vB^T  \vg_\mu 
    \end{bmatrix},
\end{align}
where we introduce another 
learning rate $\stepsize_2$ when updating $\vd$ and $\vB$.

Note that when $d_i \neq d_j$ for $i \neq j$, the $(i,j)$-th entry of the natural gradient w.r.t. $\vM$ can be expressed as
\begin{align}
   [  \vF_{MM}^{-1}\vg_{\mathrm{vecTril}(M)} \big|_{M=0} ]_{ij}=
    (\vB_k^T \vg_{S^{-1}} \vB_k)_{ij} / (d_i - d_j).
\end{align}
When $d_i = d_j$, we simply set the corresponding entry to be zero due to the
Moore-Penrose inverse.

Finally, we can re-express the above update as:
\begin{align}
    \begin{bmatrix}
        \vd_{k+1} \\
        \vB_{k+1}\\
        \vmu_{k+1} 
    \end{bmatrix} \leftarrow
        \begin{bmatrix}
        \vd_k  \odot \exp[  0 + 2 \stepsize_2  \vd_k^{-1} \odot \mathrm{diag}(\vB_k^T \vg_{S^{-1}} \vB_k) ] \\
        \vB_k \mathrm{Cayley}( \mathrm{Tril}(\vU)  - [\mathrm{Tril}(\vU)]^T ) \\
        \vmu_k - \stepsize_1 \vB_k \mathrm{Diag}(\vd_{k})^{-1} \vB_k^T  \vg_\mu\\
    \end{bmatrix}
\end{align} where the $(i,j)$ entry of $\vU$ is $[\vU]_{ij}  =  0 - \stepsize_2 [\vB_k^T \vg_{S^{-1}} \vB_k]_{ij} / (d_i - d_j)  $ for $i \neq j$ and $[\vU]_{ij}=0$ when $d_i = d_j$ thanks to the Moore-Penrose inverse.

\begin{claim}
\label{claim:cayley_map}
   The Cayley map $\mathrm{Cayley}(\vN)=(\vI+\vN)(\vI-\vN)^{-1}$ is well-defined for skew-symmetric $\vN$. Moreover, this map is injective.  
\end{claim}

\begin{proof}
To show the map is well-defined, we want to show  $(\vI-\vN)$ is non-singular.
Suppose not,  we have $\mathrm{det}(\vI-\vN)=0$. Thus, $\vN$ has an eigenvalue with $1$. By the definition of the eigenvalue, there exists a non-zero vector $\vx\neq \mathbf{0}$ so that $\vN \vx = \vx$.
Notice that
Given that $\vN$ is skew-symmetric, we have
\begin{align}
\vN + \vN^T = \mathbf{0}
\end{align} and
\begin{align}
  0 =  \vx^T (\vN+\vN^T) \vx = \vx^T (\vN \vx) + (\vx^T \vN^T) \vx = \vx^T \vx + \vx^T \vx 
\end{align} 
The above expression implies  $\vx=\mathbf{0}$, which is a contradiction.
Thus, $\mathrm{det}(\vI-\vN)\neq 0$ and $(\vI-\vN)$ is non-singular. 

Now, we show that the Cayley map is injective if $\vN$ is skew-symmetric. 
Let $\vQ = \mathrm{Cayley}(\vN)$. 
We first assume $(\vQ+\vI)$ is non-singular, and then we prove it. Given $(\vQ+\vI)$ is non-singular, we have 
\begin{align*}
   \vQ (\vI - \vN ) = (\vI+\vN) \iff \vQ -\vI = (\vQ +\vI) \vN \iff (\vQ+\vI)^{-1}(\vQ-\vI) = \vN,
\end{align*} This implies the map is injective and its inverse is 
\begin{align}
\vN=\mathrm{Cayley}^{-1}(\vQ) :=(\vQ+\vI)^{-1}(\vQ-\vI) \label{eq:inv_cayley}
\end{align}

Now, we show that $(\vQ+\vI)$ is non-singular. We use proof by contradiction.
If not, there exists a non-zero vector $\vv$ \citep{cayley-notes} so that
\begin{align}
    \vQ \vv = - \vv &\iff (\vI +\vN) (\vI-\vN)^{-1} \vv = -\vv\\
    &\iff (\vI-\vN)^{-1} (\vI+\vN)\vv = -\vv\\
    &\iff (\vI+\vN)\vv = -(\vI-\vN)\vv \\
    &\iff \vv = -\vv,\, (\text{another contradiction since } \vv \neq 0) 
\end{align} where we use the following identity in the second step in the above expression.
\begin{align}
   (\vI +\vN )(\vI-\vN)^{-1} &= -( -2\vI + \vI -\vN) (\vI-\vN)^{-1} =2 (\vI-\vN)^{-1} - \vI\\
   &= (\vI-\vN)^{-1}(2\vI - (\vI-\vN)) = (\vI-\vN)^{-1} (\vI+\vN)
\end{align}

\end{proof}

\end{document}